%% file: acl_latex.tex
\documentclass[11pt]{article}

\usepackage[preprint]{acl}

\usepackage{times}
\usepackage{latexsym}

\usepackage[T1]{fontenc}

\usepackage[utf8]{inputenc}

\usepackage{microtype}

\usepackage{inconsolata}

\usepackage{graphicx}
\input{preamble}

\newcommand{\papertitle}{What Makes a Good Unmasking Policy? A Controlled Study of Learned Unmasking in Diffusion Language Models}
\title{\papertitle}

\author{
  \textbf{Jinya Sakurai\textsuperscript{1~2}}\quad\quad
  \textbf{Xun Xu\textsuperscript{2~3}}\quad\quad
  \textbf{Jaehong Yoon\textsuperscript{1}{\hypersetup{linkcolor=black}\thanks{Corresponding author.}}}
\\
\\
  \textsuperscript{1}NTU Singapore \\
  \textsuperscript{2}Institute for Advanced Intelligence and Computing (IAIC), A*STAR \\
  \textsuperscript{3}Centre for Frontier AI Research (CFAR), A*STAR
}

\begin{document}
\maketitle

\input{sec/0_abstruct}

\vspace{-0.2in}
\begin{center}
  \href{https://github.com/jinya1013/ontrack}
    {\textcolor{black}{\faGithub\ Code}}
\end{center}
\vspace{-0.01in}

\input{sec/1_introduction}
\input{sec/2_related_work}
\input{sec/3_analysis}
\input{sec/4_method}
\input{sec/5_experiments}
\input{sec/6_conclusion}

\input{sec/7_limitation}

\input{sec/8_ethical_consideration}

\clearpage
\bibliography{custom}
\clearpage
\appendix
\twocolumn[{
\begin{center}
  {\Large\bfseries\papertitle\par}
  \vspace{0.5em}
  {\Large Supplementary Material\par}
\end{center}
\vspace{1em}
}]
\input{sec_app/a_method_details}
\input{sec_app/b_controlled_study_details}
\input{sec_app/c_experimental_details}
\input{sec_app/d_numerical_results}

\end{document}

%% file: preamble.tex
\usepackage{graphicx}
\usepackage[table]{xcolor}
\usepackage{booktabs}
\usepackage{multirow}
\usepackage{array}
\usepackage{tabularx}
\usepackage{placeins}
\usepackage{wrapfig}

\usepackage{amsmath}
\usepackage{amssymb}
\usepackage{amsfonts}
\usepackage{amsthm}
\usepackage{bm}
\usepackage{bbm}
\usepackage{mathtools}

\usepackage{algorithm}
\usepackage{algorithmic}

\usepackage{pifont}
\usepackage[most]{tcolorbox}
\usepackage{caption}
\usepackage{tikz}
\usepackage{enumitem}
\usepackage{hyperref}
\usepackage{cleveref}
\usepackage{fontawesome5}
\usepackage{simpleicons}
\usepackage{balance}

\setlist[itemize]{leftmargin=2em}
\setlist[enumerate]{leftmargin=2em}

\hypersetup{
  colorlinks=true,
  linkcolor=red,
  citecolor=blue,
  urlcolor=magenta
}

\definecolor{citecolor}{RGB}{100,100,200}
\definecolor{lrssyellow}{RGB}{255,245,204}
\definecolor{lightgraytext}{gray}{0.55}

\definecolor{owtpanel}{HTML}{F3F6F8}
\definecolor{owtgray}{HTML}{68777F}
\newcolumntype{Y}{>{\raggedright\arraybackslash}X}

\theoremstyle{definition}

\theoremstyle{remark}

\def\eqref#1{equation~\ref{#1}}

\def\1{\bm{1}}

\DeclareMathAlphabet{\mathsfit}{\encodingdefault}{\sfdefault}{m}{sl}
\SetMathAlphabet{\mathsfit}{bold}{\encodingdefault}{\sfdefault}{bx}{n}

\makeatletter
\renewcommand{\maketag@@@}[1]{\hbox{\m@th\normalsize\normalfont#1}}
\makeatother

\definecolor{bestgold}{RGB}{255,243,205}
\definecolor{secondsilver}{RGB}{232,232,232}

\usepackage[most]{tcolorbox}
\newtcolorbox{findingbox}[1]{
    enhanced, breakable,
    frame hidden, boxrule=0pt, sharp corners,
    colback=black!3,
    borderline west={1.2pt}{0pt}{black!45},
    left=6pt, right=6pt, top=4pt, bottom=4pt,
    before skip=6pt, after skip=6pt,
    before upper={\textbf{Finding #1.}~},
}

%% file: sec/0_abstruct.tex
\begin{abstract}
Masked diffusion language models generate tokens in parallel,
but their efficiency depends on which predictions are committed.
Learned unmasking policies improve the speed--accuracy trade-off,
yet differ in supervision, inputs, and capacity, obscuring which
choices matter.
We view supervised policies as distilling a decoding procedure
through a teacher and examine independently what the teacher
labels as committable, on which states the policy imitates it, what the
policy observes, and how much the number of function evaluations (NFE)
must decrease to offset policy overhead.
We find that labeling predictions matching the final output is overly
permissive at low NFE, and hidden states are weak alone but can
complement confidence.
About a 10\% NFE reduction suffices to offset policy
overhead for policies up to 7.5M parameters across tested settings.
We therefore train a policy on trajectory-contraction targets, which
accept only commitments staying on a reference trajectory, using
confidence and hidden states.
On LLaDA-8B-Instruct, the resulting 856K-parameter policy achieves the best
accuracy--NFE frontier and highest throughput on GSM8K, is competitive at moderate targets on MATH-500 and HumanEval, and largely
transfers to Dream-7B-Instruct.
\end{abstract}

%% file: sec/1_introduction.tex
\section{Introduction}
\label{sec:introduction}

Building on discrete diffusion models
\citep{hoogeboom2021argmax,austin2021structured,
campbell2022a,lou2024discrete},
masked diffusion language models (dLLMs)
\citep{sahoo2024simple,shi2024simplified,nie2025llada,ye2025dream,sakurai2026simplex}
provide an alternative to the left-to-right generation of autoregressive
models \citep{brown2020gpt3,grattafiori2024llama3herdmodels,qwen2025qwen25technicalreport} by iteratively refining partially masked sequences and
committing multiple tokens in parallel. Their efficiency, however, depends strongly on the unmasking
policy, which decides which positions to commit at each step. Conservative policies require many denoising evaluations, whereas
unmasking many tokens at once can produce mutually inconsistent tokens,
as they are predicted independently given the same state.
Such errors persist, since committed tokens are not revised later
\citep{wu2026fastdllm,jazbec2026learning,bao2026learning,xu2026scheduling}. Recent work therefore adapts the number of committed tokens to the decoding
state.
Training-free methods do so using confidence \citep{wu2026fastdllm} or
entropy-based dependence control \citep{ben-hamu2025accelerated}, while
learned policies are trained from data
\citep{jazbec2026learning,bao2026learning,xu2026scheduling,
sun2026tracelock}.

Despite this progress, learned unmasking policies differ in several design
choices at once: which positions are labeled as safe to commit, which
decoding states are used for training, and what information the policy
observes, often together with policy capacity.
For example, whether backbone hidden states help cannot be judged, since
the policies that use them also differ in size or supervision.
Prior work reports the number of function evaluations (NFE) and
throughput, but the NFE reduction needed to offset policy overhead
across policy sizes, caching, and prompt lengths remains unclear.

We therefore revisit learned unmasking policies from a distillation
perspective.
Existing supervised policies \citep{bao2026learning,sun2026tracelock} can
be viewed as distilling a fixed decoding procedure: the procedure is run on
training prompts to obtain reference trajectories, a teacher derived from
them labels which positions may be committed, and the policy imitates this
teacher.
This view separates what the teacher labels, on which states the policy
imitates it, and what the policy observes; we vary one at a time,
comparing choices from prior work with new candidates.
First, existing teachers label a prediction as committable whenever it
matches the final output.
A correct token, however, is not necessarily safe to commit early, since
its commitment changes the input to subsequent steps.
We therefore introduce trajectory contraction, which labels a commitment
as valid only when it directly reaches a later state of the reference
trajectory, and compare it jointly with the choice of training states:
reference trajectories or those induced by the labels.
Second, as policy inputs, we compare confidence, probability margins,
hidden representations, and confidence combined with hidden
representations.
Third, we measure both NFE and end-to-end throughput in tokens per second
(TPS), and profile the policy overhead.

Our study shows that matching the final output is an overly permissive
label in the low-NFE regime, with 28.7\% of such positives leaving the
reference trajectory when committed early, and that hidden
representations are weak alone but can complement confidence.
It also shows that, across the tested settings, a policy of up to 7.5M
parameters pays off in throughput once it reduces NFE by more than about
10\%.
Based on these findings, we train a policy with trajectory-contraction
targets on reference trajectories, taking confidence and hidden
representations as input; it has only 856K parameters, about 0.01\% of
the backbone.

On LLaDA-8B-Instruct \citep{nie2025llada}, our policy achieves the best accuracy--NFE frontier
on GSM8K \cite{cobbe2021gsm8k}, where it also attains the highest TPS at every accuracy target.
On MATH-500 \citep{hendrycks2021math,lightman2024lets} and HumanEval \citep{chen2021humaneval},
it reaches moderate accuracy targets with the fewest NFE and the highest TPS,
whereas training-free methods are more efficient at the highest targets,
and on MBPP \cite{austin2021mbpp} it performs comparably to the strongest baselines.
Its advantage largely carries over to Dream-7B-Instruct \citep{ye2025dream} and, on
mathematical reasoning, to unseen generation lengths.

Our contributions are:
\hspace{-1em}
\begin{itemize}[leftmargin=1.4em, itemsep=-0.5em]
    \item We show that matching the final output is an overly permissive
    commitment label in the low-NFE regime, and introduce trajectory contraction, which improves
    the accuracy--NFE frontier, with a further out-of-domain gain from supervision on reference
    trajectories.
    \item We show that backbone hidden states alone are a weak input, but
    can improve the frontier at high accuracy targets when combined with
    token confidence, and build on this an 856K-parameter policy.
    \item By profiling per-step latency, we show that policies of up to
    7.5M parameters pay off in throughput once they reduce NFE
    by more than about 10\% in the tested settings, and evaluate our policy across tasks,
    backbones, and generation lengths.
\end{itemize}

%% file: sec/2_related_work.tex
\section{Related Work}
\label{sec:related_work}

\noindent \textbf{Efficient unmasking.}
The efficiency and quality of dLLMs depend strongly on the order and degree of parallelism used for unmasking \citep{kim2025train,chen2026optimal}. Training-free approaches exploit predictive uncertainty, decoding dynamics, or dependencies among candidate positions to determine which tokens can be committed in parallel \citep{ben-hamu2025accelerated,wei2026slowfast,luxembourg2026plan,lee2026lookahead,kim2026dapd}. Fast-dLLM is particularly relevant to our setting: it uses confidence-thresholded parallel decoding to reduce denoising steps while keeping the underlying dLLM frozen \citep{wu2026fastdllm}. Despite substantial progress, these methods prescribe the commitment rule at inference time rather than learning it from decoding experience.

\noindent \textbf{Learning unmasking policies.}
Recent work instead learns where or when to commit tokens, using reinforcement learning, supervised oracle signals, ranking objectives, or learned dependency models \citep{hong2026improving,xu2026scheduling,asano2026wheretounmask,ringel2026dependency}. LUP learns a lightweight Transformer policy from confidence-based features using reinforcement learning \citep{jazbec2026learning}. More directly related to our supervision setting, Learn2PD trains a lightweight filter from oracle token-correctness signals, whereas TraceLock derives commitment labels retrospectively from completed decoding traces and exploits richer hidden-state representations \citep{bao2026learning,sun2026tracelock}. We study how to construct supervised commitment targets directly from Fast-dLLM trajectories and how to incorporate richer token-level information without sacrificing the efficiency of lightweight confidence-based policies.

%% file: sec/3_analysis.tex
\begin{table*}[t]
    \centering
    \small
    \begin{tabularx}{\textwidth}{l r l X l}
        \toprule
        Method & \#Params & Policy input & Commitment target & Training states \\
        \midrule
        Learn2PD \citep{bao2026learning}
            & 2.1K
            & confidence
            & terminal agreement
            & teacher-induced \\
        TraceLock \citep{sun2026tracelock}
            & 7.5M$^\dagger$
            & hidden states
            & terminal agreement
            & reference \\
        LUP-conf \citep{jazbec2026learning}
            & 300K
            & confidence
            & reinforcement learning
            & on-policy rollouts \\
        LUP-hidden \citep{jazbec2026learning}
            & 300M
            & hidden states
            & reinforcement learning
            & on-policy rollouts \\
        \midrule
        Candidates studied
            & 331K--856K
            & $c$, $h$, $[c,m]$, $[c,h]^\ast$
            & terminal agreement, contraction$^\ast$
            & reference, teacher-induced \\
        \bottomrule
    \end{tabularx}
    \caption{
    Design choices of existing learned unmasking policies and the candidates
    compared in our study.
    Existing methods differ simultaneously in their policy inputs, policy
    capacity, commitment targets, and training states.
    $^\ast$Candidates introduced in this work.
    $^\dagger$Estimated from the released configuration; not reported in the original paper.
    }
    \label{tab:existing_designs}
\end{table*}

\section{Revisiting Unmasking Policy Learning}
\label{sec:revisiting}

\subsection{Problem Formulation}
\label{sec:problem}

A masked dLLM generates a sequence $x=(x_1,\ldots,x_L)$ through partially
masked states $\tau=(s_0,\ldots,s_T)$.
At each step $t$, the model produces token distributions
$p_\theta(\cdot\mid s_t)$ for the masked positions $\mathcal{M}(s_t)$, and an
unmasking policy selects the positions to commit,
$A_t=\pi_\phi(z_t)\subseteq\mathcal{M}(s_t)$, where $z_t$ is the information
available to the policy.
The decoder then transitions to $s_{t+1}=\mathcal{C}(s_t,A_t)$, where
$\mathcal{C}$ fills the selected positions with their greedy predictions, so
the trajectory is deterministic given the policy decisions.
Committing more positions per step reduces NFE, but premature commitments may
degrade generation quality; a good policy achieves a favorable trade-off
between the two.

\subsection{Design Space and Protocol}
\label{sec:controlled_setup}

We study supervised policies trained offline on a frozen backbone, such as
Learn2PD \citep{bao2026learning} and TraceLock \citep{sun2026tracelock}.
These policies distill a fixed decoding procedure: the procedure is run on
training prompts to obtain \emph{reference trajectories}, from which each
masked position at each state is labeled with a binary \emph{commitment
target} indicating whether it may be committed, and the policy is trained to
predict these targets.
Building such a policy requires deciding what counts as a correct commitment
(the target), on which states the policy learns it (the training states;
\Cref{sec:supervision}), and what the policy observes (the input;
\Cref{sec:policy_input}).
Moreover, the policy is evaluated at every step, so its cost may offset the
reduction in NFE (\Cref{sec:nfe_tps}).

{Existing methods vary simultaneously across multiple design axes, making it difficult to isolate the effect of individual design choices from their reported results.
For example, LUP-hidden underperforms LUP-conf despite roughly
$1000\times$ more parameters \citep{jazbec2026learning} (\Cref{tab:existing_designs}).
We therefore compare existing and newly introduced candidates for each design axis while holding the others fixed. We evaluate the commitment target and training states jointly in a $2\times2$ design to account for their interaction during teacher rollouts.}

\noindent\textbf{Setup.} We use LLaDA-8B-Instruct \citep{nie2025llada} as the frozen backbone.
Following \citet{jazbec2026learning}, all policies are trained on a mixture
of the GSM8K \citep{cobbe2021gsm8k} and MATH \citep{hendrycks2021math}
training sets, with reference trajectories generated by Fast-dLLM
\citep{wu2026fastdllm} at confidence threshold $0.9$.
Unless otherwise specified, we use generation length $L=256$ and block
length $32$, and evaluate zero-shot.
We take the accuracy of one-by-one confidence-based decoding
\citep{nie2025llada} as the reference and define target {accuracy levels} at
$98\%$, $95\%$, and $90\%$ of it.
For each method, we sweep its commitment threshold over a fixed grid
(\Cref{app:param_grids}) and compare the resulting accuracy--NFE
trade-off against these targets.
In this section, we evaluate on GSM8K (in-domain) and MBPP (out-of-domain) \citep{austin2021mbpp}, and all candidates share the same
architecture, optimizer, and training schedule
(\Cref{app:controlled_setup,app:training_configuration}).

\begin{figure*}[t]
    \centering
    \includegraphics[width=\linewidth]{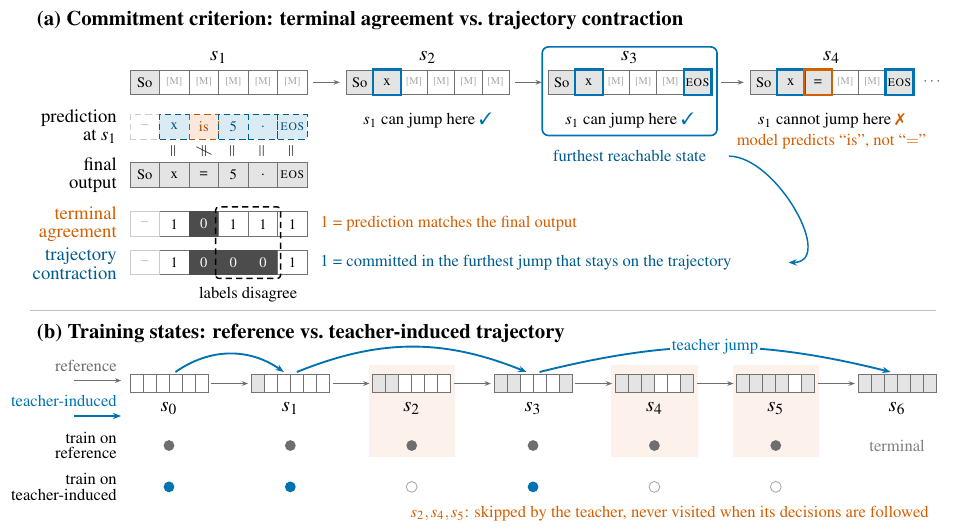}
    \vspace{-0.2in}
    \caption{
    Design choices in supervising unmasking policies.
    \textbf{(a)} Terminal agreement labels every prediction matching the final
    output, whereas trajectory contraction labels only positions in the furthest
    jump that stays on the reference trajectory; they disagree when a correct
    token is committed before a mispredicted one.
    \textbf{(b)} Following the teacher skips $s_2$, $s_4$, and $s_5$, which
    reference-trajectory supervision includes but teacher-induced supervision omits.
    }
    \label{fig:supervision_concept}
\end{figure*}

\subsection{How Should Supervision Be Constructed?}
\label{sec:supervision}

\noindent\textbf{Commitment targets.}
We consider two oracle criteria for defining supervision targets
(\Cref{fig:supervision_concept}, panel (a)).
Each oracle specifies which masked positions should be committed at
a given state.
The first, \emph{terminal agreement}, is used by both Learn2PD \citep{bao2026learning} and
TraceLock \citep{sun2026tracelock}.
It commits a position when its current prediction $\hat{x}_{t,i}$
matches the terminal token $x_i^\star$ of the reference trajectory:
$y_{t,i}^{\mathrm{tok}}=\mathbbm{1}[\hat{x}_{t,i}=x_i^\star]$
(\Cref{app:target_construction}).
However, matching the terminal output does not guarantee that early
commitment is safe: committing a token changes the input to subsequent
evaluations and may alter predictions at other masked positions.
We therefore introduce \emph{trajectory contraction} (TC), which
permits a commitment only if it directly reaches a state visited
later in the reference trajectory.
Let $A_{t\rightarrow u}=\mathcal{M}(s_t)\setminus\mathcal{M}(s_u)$
denote the positions committed between $s_t$ and a future state
$s_u$ of the reference trajectory.
We consider the commitment of this set valid if
\begin{equation}
    \mathcal{C}(s_t,A_{t\rightarrow u}) = s_u.
    \label{eq:trajectory_contraction}
\end{equation}
This condition allows the intermediate states to be skipped while
preserving the future reference state.
We define $u_t^\star$ as the largest $u>t$ satisfying \Cref{eq:trajectory_contraction}, and set the target $y_{t,i}^{\mathrm{tc}}=\mathbbm{1}[i\in A_{t\rightarrow u_t^\star}]$.

\noindent\textbf{Training states.}
In addition to the commitment criterion, we consider two ways of
constructing the states used for supervision
(\Cref{fig:supervision_concept}, panel (b)).
With \emph{reference trajectories}, we collect training states from
the fixed decoding procedure and apply the oracle at each state,
including those the oracle would otherwise skip.
With \emph{teacher-induced trajectories}, we instead generate training
states by repeatedly applying the oracle's commitment decisions,
so that supervision is provided only at the states it visits.
TraceLock uses reference trajectories, whereas Learn2PD uses
teacher-induced trajectories, referring to its oracle as the teacher.

\begin{figure}[t]
    \centering
    \includegraphics[width=\linewidth]{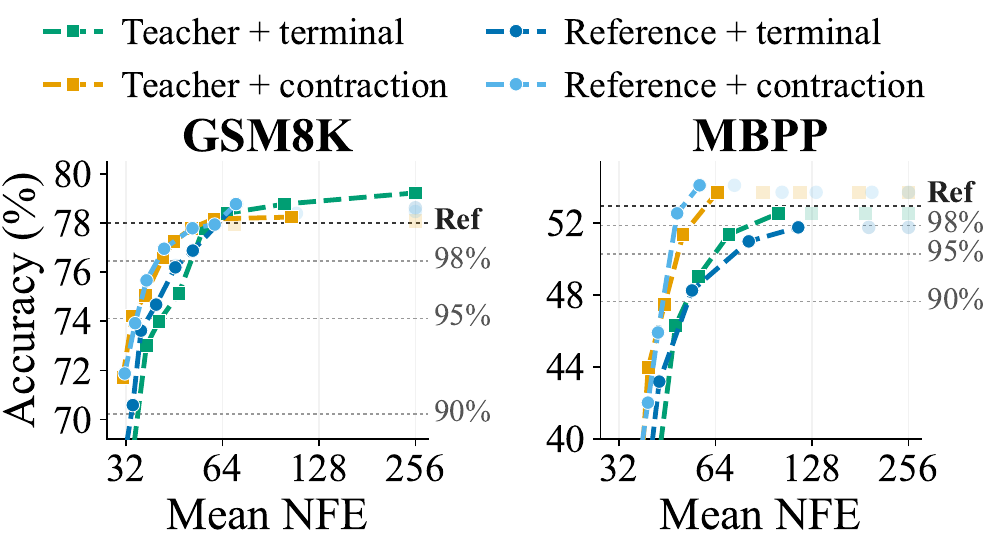}
    \vspace{-0.25in}
    \caption{
    Accuracy--NFE frontiers for the $2\times2$ supervision designs.
    Faded markers denote dominated operating points.
    }
    \label{fig:teacher_signal}
    \vspace{-0.15in}
\end{figure}

\noindent\textbf{Comparison.}
We compare the four combinations of two commitment targets
(terminal agreement and trajectory contraction) and two training-state
constructions (reference and teacher-induced trajectories),
with the policy input fixed (\Cref{app:controlled_setup}).
As shown in \Cref{fig:teacher_signal}, replacing terminal agreement with
trajectory contraction improves the frontier under both training-state
constructions.
On GSM8K, this change accounts for most of the gain, and both
constructions perform comparably under contraction.
On the out-of-domain MBPP, reference trajectories yield a further
improvement, reaching the 98\% target with 48.6 NFE versus 64.8 for
teacher-induced trajectories.
The gain from contraction is concentrated at low NFE; at the highest
targets, terminal agreement can be more accurate
(\Cref{app:controlled_extra}).

Two observations help interpret these results.
First, terminal agreement is more permissive: 28.7\% of its positive
labels fall outside the contraction set (\Cref{app:target_construction}).
These are correct predictions that follow a misprediction in the
reference commitment order, so committing them early leads to a state
the reference trajectory never visits.
Second, teacher-induced trajectories retain only 34\% (contraction) and
25\% (terminal agreement) of the reference states
(\Cref{app:training_trajectory}).
This reduced coverage may explain why reference trajectories
transfer better to MBPP, although it has little effect on GSM8K.

\begin{findingbox}{1}
Trajectory contraction yields a better accuracy--NFE frontier than
terminal agreement, mainly in the low-NFE regime; training on reference
trajectories gives a further gain out of domain.
\end{findingbox}

\subsection{What Information Should the Policy Use?}
\label{sec:policy_input}

\begin{figure}[t]
    \centering
    \includegraphics[width=\linewidth]{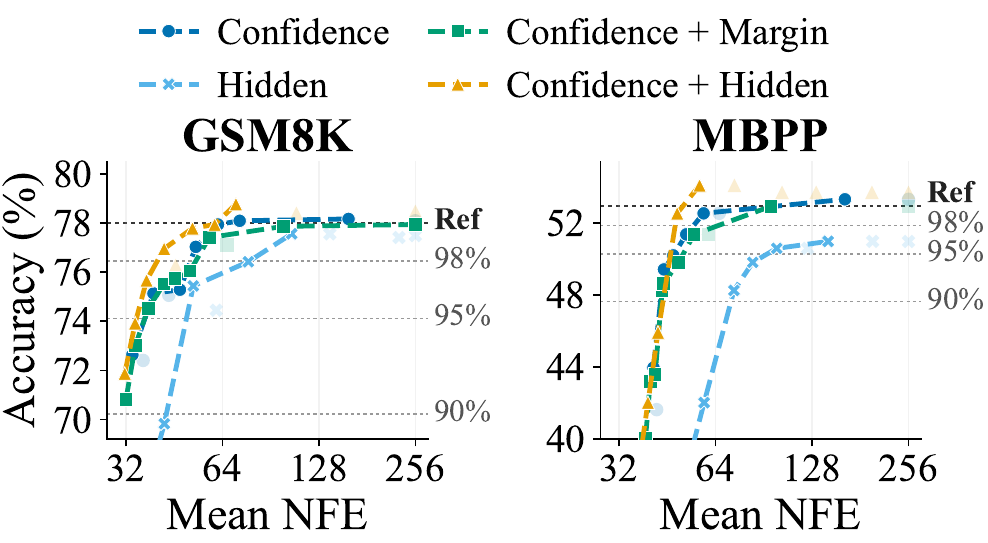}
    \vspace{-0.25in}
    \caption{
    Accuracy--NFE frontiers for different inputs.
    }
    \label{fig:input_ablation}
    \vspace{-0.15in}
\end{figure}

\noindent\textbf{Input Candidates.}
Existing policies use token confidence (Learn2PD, LUP-conf) or backbone
hidden representations (TraceLock, LUP-hidden).
Beyond confidence, uncertainty statistics such as the top-two probability
margin \citep{kim2025train} and entropy \citep{xu2026scheduling} are also
used to decide which tokens to unmask.
For each masked position $i$ at step $t$, we consider the token confidence
$c_{t,i}=\max_v p_\theta(x_i=v\mid s_t)$, the top-two probability margin
$\Delta_{t,i}$ defined as the difference between the largest and
second-largest token probabilities, and the backbone hidden
representation $h_{t,i}$.
We compare four input configurations:
\begin{equation}
    z_{t,i}
    \in
    \left\{
        c_{t,i},\;
        h_{t,i},\;
        [c_{t,i},\Delta_{t,i}],\;
        [c_{t,i},h_{t,i}]
    \right\}.
\end{equation}
Each input is projected to a fixed hidden dimension and processed
by the policy network $f_\phi$, which predicts a commitment score
for each masked position.
To isolate the effect of the input from policy capacity
(\Cref{sec:controlled_setup}), we keep the policy architecture fixed
across configurations, varying only the input projection.

\noindent\textbf{Comparison.}
As shown in \Cref{fig:input_ablation}, confidence provides a strong
low-dimensional signal, whereas hidden states alone require substantially
more NFEs to reach the same accuracy on both tasks.
Adding the margin yields no consistent gain.
Combining confidence with hidden states reaches the 98\% and 95\% targets
with the fewest NFE on both tasks, e.g., 42.1 versus 52.9 NFE for
confidence alone at the 98\% target on GSM8K.
At the 90\% target, however, confidence alone is comparable on GSM8K and
slightly more efficient on MBPP.
On MATH-500 and HumanEval, the benefit is less consistent, and adding the
margin is competitive or better at the highest targets
(\Cref{app:controlled_extra}).
These results suggest that hidden states can complement confidence;
we hypothesize that their high dimensionality makes it difficult for a
lightweight policy to exploit them alone.
This gain comes at a cost: projecting the 4096-dimensional hidden state
increases the policy from 331K to 856K parameters.
\begin{findingbox}{2}
Hidden representations underperform confidence when used alone but can
complement it, mainly reducing the NFE needed to reach high accuracy
targets; the gain varies across tasks.
\end{findingbox}

\subsection{Do NFE Gains Translate to Speedups?}
\label{sec:nfe_tps}

The comparisons above measure decoding efficiency by NFE.
While NFE captures the number of backbone evaluations, it does not
account for the computational overhead of the policy itself.
This trade-off depends on policy size (\Cref{tab:existing_designs}),
including the 331K-to-856K increase from $[c,h]$, and on backbone
latency, which varies with prompt length and caching
\citep{wu2026fastdllm}.
We therefore profile per-step latency across these settings to determine
how much NFE reduction is needed to improve throughput
(\Cref{tab:step_time}; details in \Cref{app:step_time}).
Besides our 331K and 856K policies, we include randomly initialized
policies of 7.5M and 300M parameters, approximately matching TraceLock and
LUP-hidden.

\input{tab/step_time}

The key result is that the break-even NFE reduction remains small for
lightweight policies.
Up to 7.5M parameters, the policy adds a nearly constant cost of about
1.5\,ms per step, at most 10\% across all settings.
Thus, reducing NFE by about 10\% is sufficient to offset the policy
overhead, regardless of prompt length or caching.
Even a 300M policy, with up to 39\% overhead, requires only about
28\% fewer NFEs to break even.
Although prefix caching reduces backbone latency by 73\% at 2048 prompt
tokens but only 5\% at 128 tokens, the break-even threshold for
policies up to 7.5M parameters remains around 10\%.
We evaluate the resulting throughput gains end to end in
\Cref{sec:main_results,sec:prefix_cache}.

\begin{findingbox}{3}
Policies with up to 7.5M parameters can improve throughput with
a 10\% NFE reduction across prompt lengths and caching.
\end{findingbox}

%% file: tab/step_time.tex
\begin{table}[t]
    \centering
    \small
    \setlength{\tabcolsep}{4pt}
    \begin{tabular}{rl r rrrr}
        \toprule
        & & No policy & \multicolumn{4}{c}{Policy overhead} \\
        \cmidrule(lr){4-7}
        Prompt & Cache & (ms/step) & 331K & 856K & 7.5M & 300M \\
        \midrule
        128  & None   & 16.9 & +10\% & +8\% & +9\% & +38\% \\
             & Prefix & 16.0 & +8\% & +8\% & +9\% & +39\% \\
        \midrule
        512  & None   & 28.0 & +5\% & +6\% & +6\% & +23\% \\
             & Prefix & 16.5 & +7\% & +10\% & +8\% & +37\% \\
        \midrule
        1024 & None   & 45.6 & +3\% & +3\% & +4\% & +15\% \\
             & Prefix & 18.6 & +8\% & +7\% & +8\% & +34\% \\
        \midrule
        2048 & None   & 85.1 & +2\% & +2\% & +2\% & +8\% \\
             & Prefix & 23.4 & +6\% & +6\% & +6\% & +27\% \\
        \bottomrule
    \end{tabular}
    \caption{
        Per-step latency on LLaDA-8B-Instruct without a policy, and the
        relative overhead of policies of different sizes.
        Absolute latencies are reported in \Cref{tab:step_time_full}.
    }
    \label{tab:step_time}
\end{table}

%% file: sec/4_method.tex
\section{Putting the Findings Together}
\label{sec:method}

We combine the findings into a single policy, which we call \emph{OnTrack}: Finding~1 determines its
supervision (\Cref{sec:policy_training}), Finding~2 its input
(\Cref{sec:policy_architecture}), and Finding~3 keeps its overhead small.

\subsection{Policy Architecture}
\label{sec:policy_architecture}

Following \citet{jazbec2026learning}, we use a DiT-style policy network for
unmasking decisions.
For each masked position $i$ in the decoding step $t$, the confidence
$c_{t,i}$ and hidden representation $h_{t,i}$ are projected onto a shared
$d$-dimensional space and additively combined:
\begin{equation}
    z_{t,i}
    =
    W_c c_{t,i}
    +
    W_h h_{t,i}.
    \label{eq:policy_input}
\end{equation}
Let $z_t=(z_{t,1},\ldots,z_{t,L})$ denote the resulting sequence.
Conditioned on the mask state $m_t$ and decoding step $t$, the policy produces
a sequence of commitment logits
\begin{equation}
    \ell_t
    =
    f_\phi(z_t;m_t,t),
    \qquad
    q_{t,i}
    =
    \sigma(\ell_{t,i}).
    \label{eq:policy_score}
\end{equation}
Both projections and $f_\phi$ are trained jointly.
With $d=128$, the policy has 856K parameters, about 0.01\% of the backbone (Architecture details are in \Cref{app:policy_architecture}).
At each step, the policy commits

\begin{equation}
    A_t
    =
    \left\{
        i\in\mathcal{E}_t:
        q_{t,i}>\gamma
    \right\},
\end{equation}

where $\mathcal{E}_t$ is the set of masked positions in the current block
and $\gamma$ is an inference-time commitment threshold
(\Cref{app:decoding}).

\subsection{Policy Training}
\label{sec:policy_training}

We generate reference trajectories as described in
\Cref{sec:controlled_setup} and retain those whose final answer is correct.
For each reference state $s_t$, the target is the trajectory-contraction
label $y_{t,i}=y^{\mathrm{tc}}_{t,i}$ (\Cref{sec:supervision}), which marks
the positions in the furthest valid contraction $A_{t\rightarrow u_t^\star}$. We optimize binary cross-entropy over eligible positions,
\begin{align}
    \mathcal{L}
    = -\!\sum_{t,\ i\in\mathcal{E}_t}
    \left[
        y_{t,i}\log q_{t,i}
        +
        (1-y_{t,i})\log(1-q_{t,i})
    \right].
    \notag
\end{align}
The complete target-construction procedure is provided in
\Cref{app:contraction_algorithm}.

%% file: sec/5_experiments.tex
\section{Experiments}
\label{sec:experiments}

\begin{figure*}[t]
    \centering
    \includegraphics[width=\linewidth]{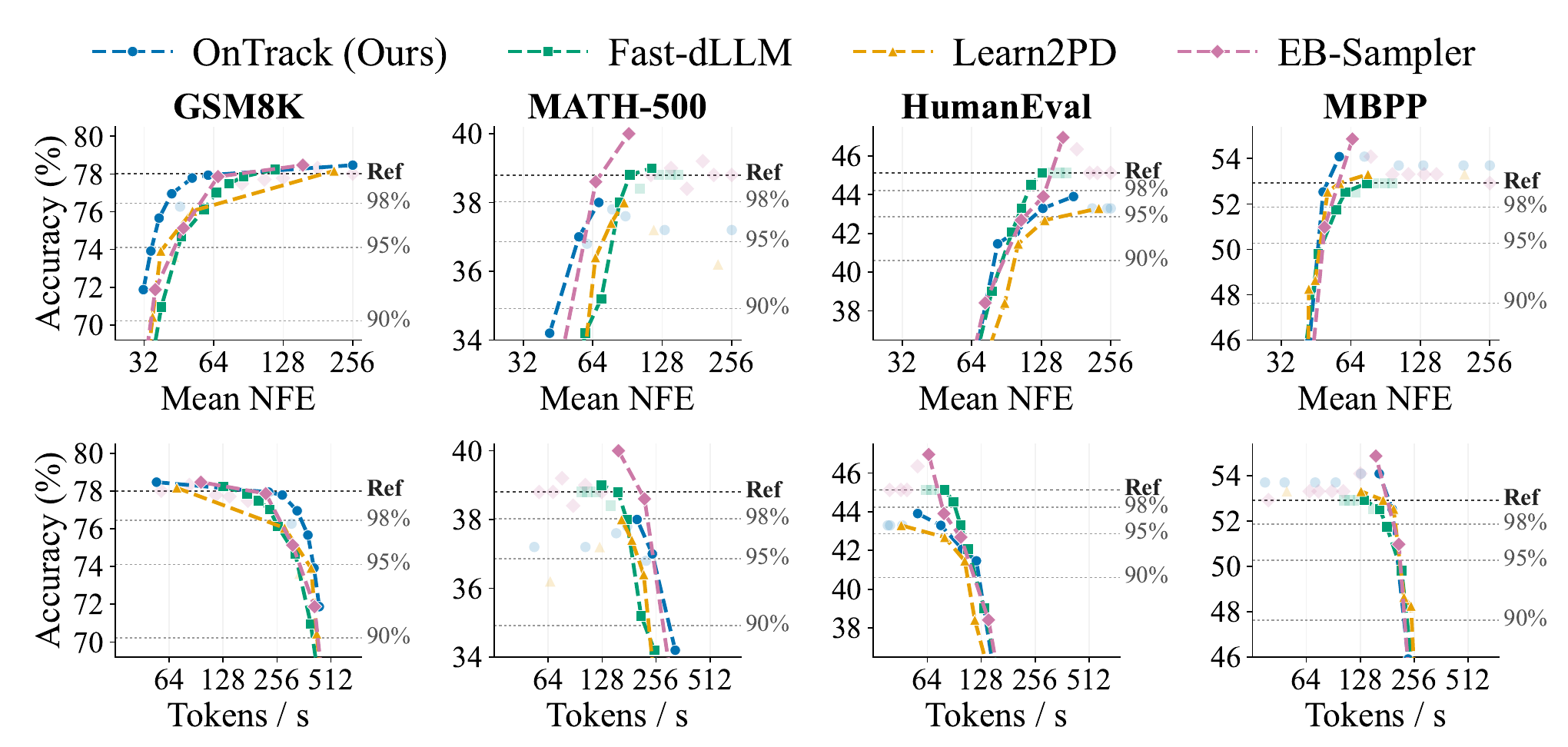}
    \vspace{-0.2in}
    \caption{
        Accuracy versus NFE (top) and TPS (bottom) on LLaDA-8B-Instruct.
        Dotted lines mark the reference accuracy and the 98\%, 95\%, and
        90\% targets.
        Numerical results are provided in \Cref{tab:results_main_math,tab:results_main_code}.
    }
    \label{fig:main_results}
\end{figure*}

\subsection{Experimental Setup}
\label{sec:experimental_setup}

\noindent\textbf{Benchmarks and backbones.}
We follow the setup of \Cref{sec:controlled_setup} and extend the
evaluation to MATH-500 \citep{hendrycks2021math,lightman2024lets} and HumanEval
\citep{chen2021humaneval}, so that GSM8K and MATH-500 evaluate in-domain
generalization and HumanEval and MBPP evaluate out-of-domain generalization
to code.
We additionally evaluate on Dream-7B-Instruct \citep{ye2025dream}
(\Cref{sec:backbone_generalization}); implementation details
are provided in \Cref{app:experimental_details}.

\noindent\textbf{Baselines.}
We compare against two training-free methods, Fast-dLLM
\citep{wu2026fastdllm}, which commits positions whose confidence exceeds a
threshold, and EB-Sampler \citep{ben-hamu2025accelerated}, which commits
positions within an entropy budget, and against Learn2PD
\citep{bao2026learning}, a learned policy for which we use the official checkpoint. We do not include LUP \citep{jazbec2026learning} or TraceLock \citep{sun2026tracelock}, as no trained checkpoints are publicly available for either method.

\input{tab/prefix_cache_gsm8k}

\noindent\textbf{Evaluation protocol.}
We report accuracy on GSM8K and MATH-500 and Pass@1 on HumanEval and MBPP,
together with NFE and TPS, with all throughput measured on the same
hardware and configuration (\Cref{app:throughput}).

\subsection{Main Results on LLaDA}
\label{sec:main_results}

\noindent\textbf{In-domain reasoning.}
On GSM8K, our policy yields the best frontier and reaches 98\% of the
reference accuracy with the fewest NFE (\Cref{fig:main_results}, top).
On MATH-500, it reaches the 95\% target with the fewest NFE, but its accuracy saturates just below the 98\% target, whereas EB-Sampler and Fast-dLLM exceed the reference at larger NFE.

\begin{figure*}[t]
    \centering
    \includegraphics[width=\linewidth]{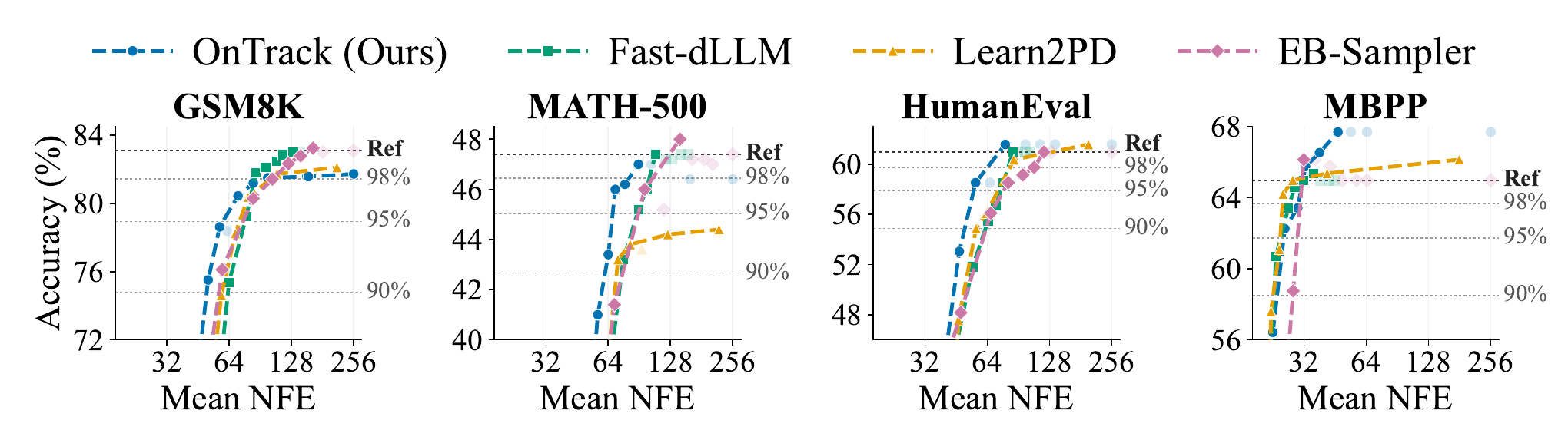}
    \vspace{-0.3in}
    \caption{
        Accuracy versus NFE on Dream-7B-Instruct.
        Dotted lines mark the reference accuracy and 98\%, 95\%, and
        90\% targets.
        Faded markers denote dominated operating points.
        Numerical results are in \Cref{tab:results_dream_math,tab:results_dream_code}.
    }
    \label{fig:dream_results}
    \vspace{-0.05in}
\end{figure*}

\begin{figure}[t]
    \centering
    \vspace{-0.05in}
    \includegraphics[width=\linewidth]{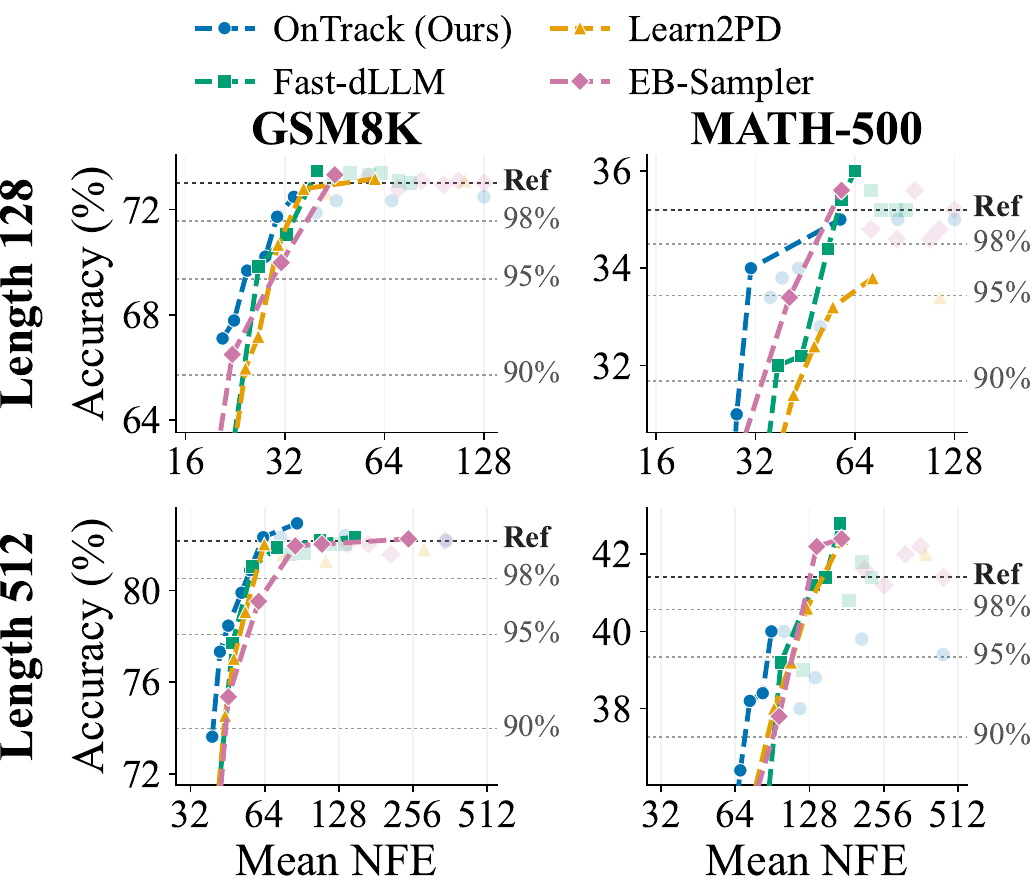}
    \vspace{-0.2in}
    \caption{
        Accuracy-NFE trade-off at $L=128$ (top) and
        $L=512$ (bottom), using a policy trained at $L=256$.
        Results on HumanEval and MBPP are provided in
        \Cref{app:length_results}; numerical results in \Cref{tab:results_len128_math,tab:results_len512_math}.
    }
    \label{fig:length_results}
    \vspace{-0.1in}
\end{figure}

\noindent\textbf{Out-of-domain code generation.}
Although the policy is trained only on mathematical reasoning, it transfers
to code.
On HumanEval, it reaches 90\% of the reference accuracy with the fewest NFE, but Fast-dLLM is more efficient at the 95\% target, and only Fast-dLLM and EB-Sampler reach the 98\% target. On MBPP, our policy reaches the 95\% and 98\% targets with the fewest NFE, whereas Learn2PD and Fast-dLLM are more efficient at the 90\% target. Paired-bootstrap confidence intervals (\Cref{tab:results_llada_ci}) show that
the NFE gains on GSM8K exceed evaluation noise against all baselines except
EB-Sampler at the 98\% target.

\noindent\textbf{Throughput.}
The NFE reductions carry over to throughput
(\Cref{fig:main_results}, bottom): at matched accuracy, our policy achieves the highest TPS on GSM8K across targets and on MATH-500 and HumanEval at moderate targets. On MBPP, it is close to the fastest baseline at the 95\% and 98\% targets but slower at the 90\% target.
Consistent with Finding~3, the 856K-parameter policy's NFE advantage outweighs the overhead of evaluating it per step.

\subsection{Compatibility with Prefix Caching}
\label{sec:prefix_cache}

We evaluate our policy and baselines with prefix caching \citep{wu2026fastdllm} on GSM8K
using zero-shot and four-shot prompts, following the latency analysis
in \Cref{sec:nfe_tps}.
To ensure consistent accuracy targets across cached and uncached decoding,
we use the uncached one-by-one baseline under the same prompting
as the reference (78.0\% zero-shot and 79.9\% four-shot).

\noindent\textbf{Fewer NFE still means faster decoding.}
The reduction in NFE carries over to caching (\Cref{tab:prefix_cache}):
with four-shot prompts, our policy remains the fastest at 98\% of the
reference accuracy.
Despite the reduced backbone latency and increased relative policy
overhead, its NFE advantage translates into higher throughput.

\noindent\textbf{Accuracy sensitivity.}
Caching reduces our policy's accuracy by up to about two percentage
points, while the baselines are less affected
(\Cref{fig:prefix_cache_accuracy}).
With zero-shot prompts, Fast-dLLM overtakes our policy at the
98\% accuracy target, although ours remains the fastest at 95\%
(\Cref{tab:prefix_cache}; see \Cref{sec:limitations}).

\subsection{Generalization Across Backbones}
\label{sec:backbone_generalization}

We next examine whether our design transfers to a different dLLM.
We apply the same procedure to Dream-7B-Instruct, training a separate policy
with the same training data, trajectory-contraction supervision, and
policy-training procedure.
Since Dream has a hidden dimension of 3584, the policy has 790K parameters.

\Cref{fig:dream_results} shows that the advantage of our policy is largely
preserved.
It yields the best frontier on HumanEval, where it exceeds the reference
accuracy with the fewest NFE, and reaches 98\% of the reference accuracy on
MATH-500 with the fewest NFE.
On GSM8K, it reaches 95\% of the reference accuracy with the fewest NFE but
saturates slightly below the reference, where Fast-dLLM and EB-Sampler
are more accurate at larger NFE.
On MBPP, all methods reach the targets within a narrow NFE range and
perform comparably.

\subsection{Generalization Across Sequence Lengths}
\label{sec:length_generalization}

We finally examine whether the learned policy generalizes across generation
lengths without retraining.
Following the length-transfer setting of \citet{jazbec2026learning}, we train
the policy with $L=256$ and evaluate the same checkpoint at $L=128$ and
$L=512$ on GSM8K and MATH-500, keeping the block length fixed at $32$.

As shown in \Cref{fig:length_results}, the policy largely retains its
advantage under both length shifts: it yields the best or comparable
frontier up to 95\% of the reference accuracy at both lengths, and at
$L=128$ it reaches 98\% on MATH-500 with the fewest NFE.
At $L=512$, it saturates below 98\% on MATH-500, where Fast-dLLM and
EB-Sampler exceed the reference at larger NFE.
When the length shift is combined with the domain shift to code, however,
the policy falls behind Fast-dLLM and EB-Sampler on HumanEval and MBPP
(\Cref{app:length_results}), indicating that it tolerates either shift
alone but not both.

%% file: tab/prefix_cache_gsm8k.tex
\begin{table}[!t]
    \centering
    \small
    \setlength{\tabcolsep}{3.5pt}
    \resizebox{0.49\textwidth}{!}{%
    \begin{tabular}{l rr rr rr rr}
    \toprule
     & \multicolumn{4}{c}{Zero-shot} & \multicolumn{4}{c}{Four-shot} \\
    \cmidrule(lr){2-5} \cmidrule(lr){6-9}
     & \multicolumn{2}{c}{98\%} & \multicolumn{2}{c}{95\%}
     & \multicolumn{2}{c}{98\%} & \multicolumn{2}{c}{95\%} \\
    \cmidrule(lr){2-3} \cmidrule(lr){4-5} \cmidrule(lr){6-7} \cmidrule(lr){8-9}
    Method & No & Pre. & No & Pre. & No & Pre. & No & Pre. \\
    \midrule
    OnTrack (Ours) & \textbf{331} & 249          & \textbf{381} & \textbf{423}
                   & \textbf{121} & \textbf{225} & \textbf{149} & 241 \\
    Fast-dLLM      & 232          & \textbf{276} & 323          & 349
                   & 98           & 193          & 122          & 234 \\
    Learn2PD       & 242          & 253          & 282          & 299
                   & 92           & 183          & 110          & \textbf{247} \\
    EB-Sampler     & 221          & 234          & 313          & 329
                   & 120          & 171          & 120          & 231 \\
    \bottomrule
    \end{tabular}}
    \caption{
        Highest measured TPS (tokens/s) on GSM8K reaching 98\% and 95\% of
        the uncached reference accuracy, without (No) and with (Pre.)
        prefix caching.
        Bold marks the best method in each column.
    }
    \label{tab:prefix_cache}
    \vspace{-0.1in}
\end{table}

%% file: sec/6_conclusion.tex
\section{Conclusion}
\label{sec:conclusion}

We revisited learned unmasking policies for dLLMs
through a controlled study of supervision, inputs, and overhead.
Trajectory-contraction targets improve the accuracy--NFE frontier mainly
at low NFE, with a further out-of-domain gain from reference
trajectories.
Hidden representations are weak alone but can complement confidence at
high accuracy targets.
Policies of up to 7.5M parameters pay off in throughput once they reduce
NFE by more than about 10\% in the tested settings.
Building on these findings, our 856K-parameter policy reaches moderate
accuracy targets with the best or comparable efficiency across two
backbones, transfers from mathematical reasoning to code, and converts
its NFE reductions into throughput gains.

%% file: sec/7_limitation.tex

\section{Limitations}
\label{sec:limitations}

\noindent \textbf{Training and evaluation scope.}
Our policy is trained only on mathematical reasoning data at a single
generation length ($L=256$) with a fixed block length.
It transfers to code at the training length and to other lengths on
mathematical tasks, but its advantage does not persist when both shifts
occur together (\Cref{sec:length_generalization}).
Training on multiple domains and generation lengths is a natural remedy
that we leave to future work.

\noindent \textbf{High-accuracy regime.}
On some tasks (MATH-500 on LLaDA and GSM8K on Dream), the accuracy of our
policy saturates slightly below the reference, even with conservative
thresholds, whereas training-free methods reach the reference at larger
NFE.
Our policy is therefore most beneficial at moderate accuracy targets, and
closing this gap may require supervision that covers more conservative
commitment behavior.

\noindent \textbf{Sensitivity to backbone features.}
Approximate inference such as prefix caching reduces the accuracy of our
policy by up to about two points, whereas confidence-based baselines are
essentially unaffected (\Cref{sec:prefix_cache}).
The drop persists even when the policy unmasks one token per step, so it
concerns which tokens are committed rather than how many.
A plausible cause is that the policy consumes backbone hidden
representations, which caching perturbs; making learned unmasking policies
robust to approximate inference is left to future work.

%% file: sec/8_ethical_consideration.tex
\section{Ethical Considerations}

\noindent\textbf{Environmental impact.}
This work aims to reduce the computational cost of inference with
diffusion language models, potentially lowering energy consumption.
However, the overall environmental impact depends on deployment scale,
as improved efficiency may encourage greater model usage.

\noindent\textbf{Potential risks.}
Faster inference may facilitate large-scale generation of misleading
or harmful content.
Our method does not modify backbone parameters and only determines
when to commit the backbone's predictions.
It therefore inherits the capabilities, biases, and failure modes
of the underlying models.
Changes in decoding may also affect output safety, which we do not
explicitly evaluate.

\noindent\textbf{Artifacts and licensing.}
All datasets (GSM8K \citep{cobbe2021gsm8k}, MATH \citep{hendrycks2021math,lightman2024lets}, HumanEval \citep{chen2021humaneval}, and MBPP \citep{austin2021mbpp}) and models
(LLaDA-8B-Instruct \citep{nie2025llada} and Dream-7B-Instruct \citep{ye2025dream}) are publicly available
and used in accordance with their licenses.
Our implementation builds on the publicly released codebase of
Learning Unmasking Policies for Diffusion Language Models
\citep{jazbec2026learning}, with modifications for our proposed
policy training and decoding methods.
We collect no new data involving human subjects.

\noindent\textbf{Use of AI assistants.}
We used AI assistants to support code development, figure preparation,
and language polishing of the manuscript.
All AI-assisted outputs were reviewed and verified by the authors,
who take full responsibility for the code, figures, and scientific content.

%% file: sec_app/a_method_details.tex
\section{Method Details}
\label{app:method}

\subsection{Policy Architecture}
\label{app:policy_architecture}

Our policy follows the DiT-style architecture of
\citet{jazbec2026learning}.
We use the token confidence $c_{t,i}$ and hidden representation $h_{t,i}$
defined in \Cref{sec:policy_input}.
The two signals are independently projected to a shared $d$-dimensional
representation:
\begin{align}
    z^{\mathrm{conf}}_{t,i}
    &= W_c c_{t,i} + b_c, \\
    z^{\mathrm{hid}}_{t,i}
    &= W_h h_{t,i} + b_h ,
\end{align}
where $W_c\in\mathbb{R}^{d\times 1}$ and
$W_h\in\mathbb{R}^{d\times d_h}$.
The hidden-state stream is added to the confidence representation:
\begin{equation}
    z_{t,i}
    =
    z^{\mathrm{conf}}_{t,i}
    +
    z^{\mathrm{hid}}_{t,i}.
    \label{eq:app_policy_input}
\end{equation}

Let $z_t=(z_{t,1},\ldots,z_{t,L})$ denote the resulting sequence and
$m_t=(m_{t,1},\ldots,m_{t,L})$ the corresponding mask state.
The policy produces commitment logits
\begin{equation}
    \ell_t
    =
    f_\phi(z_t;m_t,t),
    \qquad
    q_{t,i}
    =
    \sigma(\ell_{t,i}).
    \label{eq:app_policy_score}
\end{equation}
Following \citet{jazbec2026learning}, we use a single DiT-style
transformer block.
Token positions are encoded through RoPE~\citep{su2024roformer} in
self-attention.

The confidence projection, hidden-state projection, and policy network
$f_\phi$ are optimized jointly, while the dLLM backbone remains frozen
throughout training.
The policy produces logits for all sequence positions, while commitment is
restricted to masked positions in the current block, as described in
\Cref{app:decoding}.

\begin{table}[b]
    \centering
    \small
    \begin{tabularx}{\columnwidth}{X rr}
        \toprule
        Component & LLaDA & Dream \\
        \midrule
        Policy width $d$              & 128     & 128 \\
        DiT depth                     & 1       & 1 \\
        Attention heads               & 2       & 2 \\
        Confidence pathway + DiT      & 331,297 & 331,297 \\
        Hidden projection             & 524,416 & 458,880 \\
        Total policy parameters       & 855,713 & 790,177 \\
        \bottomrule
    \end{tabularx}
    \caption{
        Configuration of the unmasking policy for each backbone.
        The hidden projection maps the 4096-dimensional (LLaDA) or
        3584-dimensional (Dream) hidden state to $d$ dimensions.
    }
    \label{tab:policy_architecture}
\end{table}

\begin{figure}[t]
    \centering
    \includegraphics[width=\linewidth]{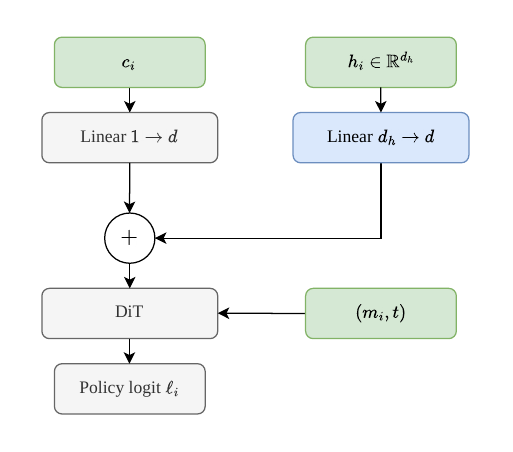}
    \caption{
        Architecture of our unmasking policy.
        Following \citet{jazbec2026learning}, confidence is projected into
        the policy representation space and processed by a DiT-style
        transformer.
        Our policy adds a projected backbone hidden representation to the
        confidence representation.
    }
    \label{fig:policy_architecture}
\end{figure}

\subsection{Trajectory-Contraction Teacher}
\label{app:contraction_algorithm}

We construct training targets from a reference trajectory
\begin{equation}
    \tau^{\mathrm{ref}}
    =
    (s_0,\ldots,s_T).
\end{equation}
For each state $s_t$ and future state $s_u$, $u>t$, we define
\begin{equation}
    A_{t\rightarrow u}
    =
    \mathcal{M}(s_t)
    \setminus
    \mathcal{M}(s_u),
\end{equation}
as the set of positions committed between the two states.
Following \Cref{sec:supervision}, the segment from $s_t$ to $s_u$ can be
contracted into a single decoding step when
\begin{equation}
    \mathcal{C}(s_t,A_{t\rightarrow u})
    =
    s_u.
    \label{eq:app_valid_contraction}
\end{equation}

For each reference state, we select the furthest future state satisfying this
criterion,
\begin{equation}
    u_t^\star
    =
    \max
    \left\{
        u>t :
        \mathcal{C}(s_t,A_{t\rightarrow u})=s_u
    \right\},
    \label{eq:app_furthest_contraction}
\end{equation}
and use the corresponding commitment set
\begin{equation}
    A_t^\star
    =
    A_{t\rightarrow u_t^\star}
\end{equation}
to define the binary teacher targets
\begin{equation}
    y_{t,i}
    =
    \mathbbm{1}\!\left[i\in A_t^\star\right].
    \label{eq:app_contraction_target}
\end{equation}
Validity is monotone in $u$: since $A_{t\rightarrow u'}\subseteq
A_{t\rightarrow u}$ for $u'<u$, if \Cref{eq:app_valid_contraction} holds for
$u$, it also holds for every $t<u'<u$.
Targets therefore require no backbone evaluations beyond those of the
reference trajectory.

We construct these targets independently for every
$s_t\in\tau^{\mathrm{ref}}$ and retain all states of the reference trajectory
for policy training.
Algorithm~\ref{alg:contraction_teacher} summarizes the procedure.

\begin{algorithm}[t]
\caption{Trajectory-contraction target construction}
\label{alg:contraction_teacher}
\begin{algorithmic}[1]
\REQUIRE Reference trajectory 
$(s_0,\ldots,s_T)$
\ENSURE Teacher targets $\{A_t^\star\}_{t=0}^{T-1}$

\FOR{$t=0,\ldots,T-1$}
    \FOR{$u=T,\ldots,t+1$}
        \STATE
        $A_{t\rightarrow u}
        \leftarrow
        \mathcal{M}(s_t)\setminus\mathcal{M}(s_u)$
        \IF{$\mathcal{C}(s_t,A_{t\rightarrow u})=s_u$}
            \STATE $A_t^\star\leftarrow A_{t\rightarrow u}$
            \STATE \textbf{break}
        \ENDIF
    \ENDFOR
\ENDFOR

\RETURN $\{A_t^\star\}_{t=0}^{T-1}$
\end{algorithmic}
\end{algorithm}

\subsection{Decoding}
\label{app:decoding}

Block diffusion partitions the generated sequence into fixed blocks
$B_1,\ldots,B_K$ and allows tokens to be committed only within the current
block $B_k$.
At each decoding step, the eligible positions are
$\mathcal{E}_t=\mathcal{M}(s_t)\cap B_k$, and the policy commits
\begin{equation}
    A_t
    =
    \left\{
        i\in\mathcal{E}_t :
        q_{t,i}>\gamma
    \right\},
\end{equation}
where $\gamma$ is the commitment threshold.
If no eligible position exceeds $\gamma$, we commit the eligible position
with the highest policy score.
The selected positions are filled with the corresponding top-1 predictions
of the frozen backbone.

%% file: sec_app/b_controlled_study_details.tex
\section{Controlled Study Details}
\label{app:controlled_study}

\subsection{Experimental Setup}
\label{app:controlled_setup}

Controlled experiments follow the training and decoding configuration in
\Cref{app:optimization,app:training_configuration} and are evaluated on
GSM8K \citep{cobbe2021gsm8k}, MATH-500 \citep{hendrycks2021math,lightman2024lets}, HumanEval \citep{chen2021humaneval}, and MBPP \citep{austin2021mbpp}.
The reference accuracy is that of one-by-one confidence-based decoding
(78.0\% on GSM8K, 38.8\% on MATH-500, 45.1\% on HumanEval, and 52.9\% on
MBPP), and the accuracy targets are 98\%, 95\%, and 90\% of it.
For each policy, we sweep the commitment threshold $\gamma$ over the grid
in \Cref{tab:param_grids} and report the frontier of non-dominated
operating points.
Each comparison varies a single factor while keeping the others fixed, as
summarized in \Cref{tab:analysis_configs}.

\begin{table}[h]
    \centering
    \small
    \begin{tabularx}{\columnwidth}{l >{\raggedright\arraybackslash}X >{\raggedright\arraybackslash}X}
        \toprule
        Section & Varied & Fixed \\
        \midrule
        \Cref{sec:supervision}
            & label, training states
            & input $[c,h]$ \\
        \Cref{sec:policy_input}
            & policy input
            & contraction, reference \\
        \bottomrule
    \end{tabularx}
    \caption{
        Factors varied and fixed in the controlled comparisons.
        All other settings follow \Cref{app:training_configuration}.
    }
    \label{tab:analysis_configs}
\end{table}

\subsection{Commitment Target Construction}
\label{app:target_construction}

Trajectory-contraction targets are constructed as described in
\Cref{app:contraction_algorithm}.

\noindent\textbf{Terminal agreement.}
We use the final token of the reference trajectory, generated by Fast-dLLM
with threshold $0.9$, as the terminal target $x_i^\star$.
This corresponds to the labeling rule shared by Learn2PD
\citep{bao2026learning} and TraceLock \citep{sun2026tracelock}, which label
a prediction by its agreement with the terminal output of a fixed decoding
procedure.
For teacher-induced trajectories, the same $x_i^\star$ is used even after
the trajectory departs from the reference.

\noindent\textbf{Agreement--contraction mismatch.}
Since a valid contraction commits only tokens that match the reference
trajectory, contraction positives are always a subset of terminal-agreement
positives.
Over all reference trajectories, 28.7\% of terminal-agreement positives
violate trajectory contraction, i.e., they are correct tokens whose
commitment at $s_t$ would leave the reference trajectory, and 56.0\% of
reference states contain at least one such violation.
The fraction is consistent across datasets (28.5\% on GSM8K and 29.1\% on
MATH).

\subsection{Training-Trajectory Construction}
\label{app:training_trajectory}

\noindent\textbf{Reference trajectories.}
Targets are computed on every state of the reference trajectory, and the
policy is trained on all of these states.

\noindent\textbf{Teacher-induced trajectories.}
Starting from $s_0$, we repeatedly apply the teacher, committing every
positively labeled position, and train on the resulting states.
Under trajectory contraction, the teacher jumps from $s_t$ to
$s_{u_t^\star}$, so the induced trajectory is a subsequence of the reference
trajectory and its targets are inherited from it.
Under terminal agreement, the induced states generally leave the reference
trajectory; we therefore run the backbone on each induced state to obtain
new predictions and recompute targets.
\Cref{tab:training_states} reports the number of training states under each
configuration.
Teacher-induced trajectories retain only a fraction of the reference states:
34.0\% under trajectory contraction and 25.4\% under terminal agreement.

\begin{table}[h]
    \centering
    \small
    \begin{tabularx}{\columnwidth}{X rr}
        \toprule
        Training states & \#States & Ratio \\
        \midrule
        Reference                   & 686,052 & 100\%  \\
        Induced (contraction)       & 233,309 & 34.0\% \\
        Induced (terminal)          & 174,390 & 25.4\% \\
        \bottomrule
    \end{tabularx}
    \caption{
        Number of training states for each training-state configuration,
        over all 9,169 reference trajectories (74.8 states per trajectory on
        average).
    }
    \label{tab:training_states}
\end{table}

\subsection{Policy-Input Study}
\label{app:policy_input}

\noindent\textbf{Inputs.}
The confidence $c_{t,i}$ and margin $\Delta_{t,i}$ are computed from the backbone's
predictive distribution at each masked position.
The hidden representation $h_{t,i}\in\mathbb{R}^{4096}$ is taken from
the final layer of the backbone.
Each input is linearly projected to $d=128$ dimensions before entering the
shared policy network $f_\phi$.

\noindent\textbf{Parameter counts.}
\Cref{tab:input_params} lists the parameter count of each policy.
All policies share the same network $f_\phi$; the differences arise only from
the input projections.

\begin{table}[h]
    \centering
    \small
    \begin{tabularx}{\columnwidth}{X r}
        \toprule
        Input & \#Params \\
        \midrule
        $c_{t,i}$              & 331,297 \\
        $h_{t,i}$              & 855,457 \\
        $[c_{t,i},\Delta_{t,i}]$    & 331,425 \\
        $[c_{t,i},h_{t,i}]$    & 855,713 \\
        \bottomrule
    \end{tabularx}
    \caption{
    Parameter counts of the policies in \Cref{sec:policy_input}.
    The DiT block accounts for 297,376 parameters and the timestep embedding
    for 33,024; the hidden-state projection adds 524,416.
    }
    \label{tab:input_params}
\end{table}

\subsection{Per-Step Latency Profiling}
\label{app:step_time}

\Cref{tab:step_time_full} reports the absolute per-step latencies
underlying \Cref{tab:step_time}.
They are obtained by profiling LLaDA-8B-Instruct on a single H200 GPU with
batch size 1, generation length $L=256$, block length 32, and commitment
threshold $0.9$, using synthetic prompts of 128 to 2048 tokens.
Each entry is the median over three decoding runs of the steady-state
per-step latency, which comprises the backbone forward pass, the policy
forward pass, and per-step bookkeeping.
Without caching, the backbone forward pass processes the full sequence;
with prefix caching, it reuses cached states, and the full forward pass
that opens each block is excluded.
Within each row, the backbone latency is the median over all policy
configurations, so differences across columns reflect only the policy and
bookkeeping.

\paragraph{Policy sizes.}
The 331K and 856K columns are the trained Conf.\ and Conf.+Hidden policies
used in our experiments.
To see how the overhead grows with policy size, we add two larger policies
whose parameter counts roughly match those of policies in prior work:
the 7.5M policy is about the size of the TraceLock \citep{sun2026tracelock} network, and the
300M policy is about the size of the LUP-hidden \citep{jazbec2026learning} network.
These two policies do not reproduce the architectures of those methods.
Both are scaled-up versions of our Conf.+Hidden policy, with widths chosen
to approximately match the target parameter counts.
The 7.5M policy uses hidden width 512, feed-forward width 2400, and 8
attention heads, with one block in each of the confidence and hidden
streams (7.52M parameters).
The 300M policy uses hidden width 2048, feed-forward width 7296, and 16
heads, with four blocks in each stream (300.3M parameters).
Both policies are randomly initialized and are used only for latency
measurement.
This does not bias the measurement, because the cost of a policy forward
pass depends on its architecture and input shape, not on its weights, and
the per-step latency is measured independently of the decisions the policy
makes.

\input{tab/step_time_full}

\subsection{Results on MATH-500 and HumanEval}
\label{app:controlled_extra}

\begin{figure}[t]
    \centering
    \includegraphics[width=\linewidth]{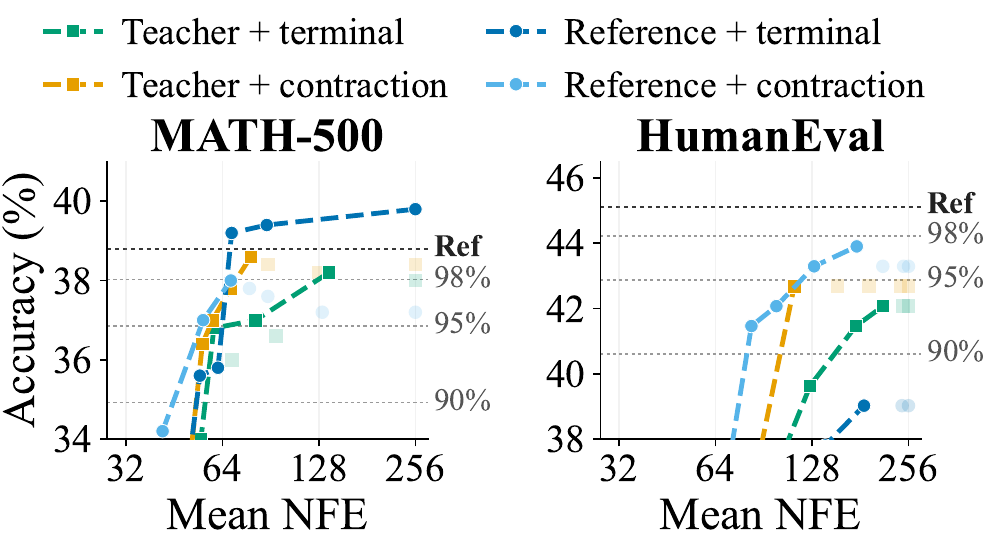}
    \caption{
    Accuracy--NFE frontiers for the $2\times2$ supervision designs on
    MATH-500 and HumanEval.
    Faded markers denote dominated operating points.
    }
    \label{fig:teacher_signal_extra}
\end{figure}

\begin{figure}[t]
    \centering
    \includegraphics[width=\linewidth]{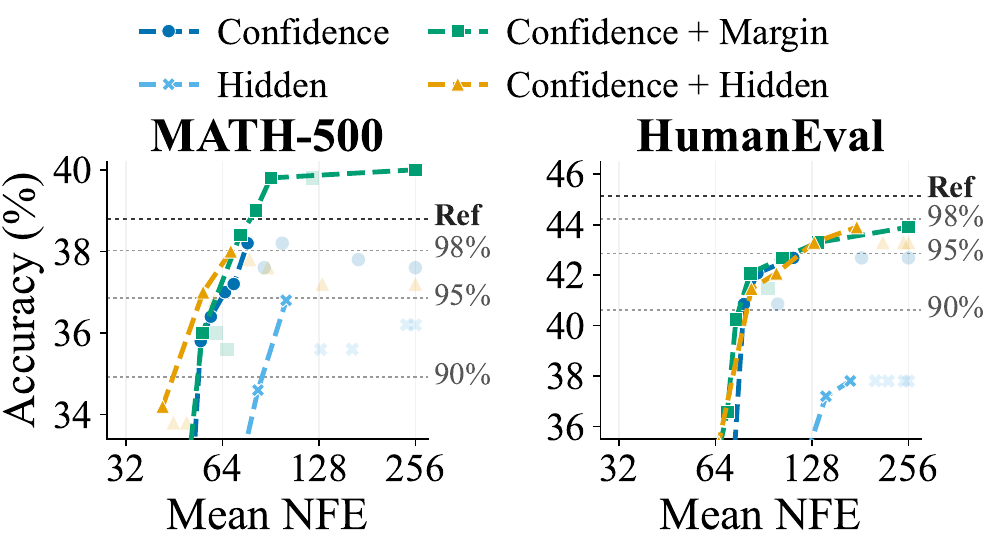}
    \caption{
    Accuracy--NFE frontiers for different policy inputs on MATH-500 and
    HumanEval.
    Faded markers denote dominated operating points.
    }
    \label{fig:input_ablation_extra}
\end{figure}

\Cref{fig:teacher_signal_extra,fig:input_ablation_extra} repeat the
comparisons of \Cref{sec:supervision,sec:policy_input} on MATH-500 and
HumanEval, with the same configurations (\Cref{tab:analysis_configs}).

\noindent\textbf{Supervision.}
On HumanEval, combining trajectory contraction with reference trajectories
yields the best frontier throughout, reaches the 90\% target with the
fewest NFE, and is the only configuration to reach the 95\% target,
whereas terminal agreement on reference trajectories remains below 90\% at
all operating points.
On MATH-500, the same combination yields the best frontier up to the 95\%
target but saturates just below 98\%, whereas the other configurations
reach 98\% at larger NFE, with terminal agreement on reference trajectories
exceeding the reference accuracy.
Trajectory contraction thus mainly improves the low-NFE regime, consistent
with Finding~1.

\noindent\textbf{Policy input.}
Hidden states alone remain below 95\% of the reference accuracy on
MATH-500 and below 90\% on HumanEval.
On MATH-500, combining confidence with hidden states yields the best
frontier up to the 95\% target but saturates just below 98\%, whereas
adding the margin reaches 98\% with the fewest NFE and the highest
accuracy, exceeding the reference.
On HumanEval, all confidence-based inputs perform comparably up to the
90\% target, but only the margin and hidden-state combinations reach
95\%, and none reaches 98\%.

%% file: tab/step_time_full.tex
\begin{table}[t]
    \centering
    \small
    \setlength{\tabcolsep}{4pt}
    \begin{tabular}{rl r rrrr}
        \toprule
        & & No policy & \multicolumn{4}{c}{With policy (ms/step)} \\
        \cmidrule(lr){4-7}
        Prompt & Cache & (ms/step) & 331K & 856K & 7.5M & 300M \\
        \midrule
        128  & None   & 16.9 & 18.5 & 18.3 & 18.4 & 23.3 \\
             & Prefix & 16.0 & 17.2 & 17.3 & 17.4 & 22.2 \\
        \midrule
        512  & None   & 28.0 & 29.4 & 29.6 & 29.6 & 34.5 \\
             & Prefix & 16.5 & 17.7 & 18.1 & 17.9 & 22.7 \\
        \midrule
        1024 & None   & 45.6 & 47.1 & 47.1 & 47.3 & 52.3 \\
             & Prefix & 18.6 & 20.0 & 19.9 & 20.0 & 24.8 \\
        \midrule
        2048 & None   & 85.1 & 86.7 & 86.8 & 86.9 & 92.2 \\
             & Prefix & 23.4 & 24.8 & 24.7 & 24.7 & 29.6 \\
        \bottomrule
    \end{tabular}
    \caption{
        Per-step latency (ms) on LLaDA-8B-Instruct with policies of different sizes
        (H200, generation length 256, block length 32, batch size 1).
        Each step is the backbone forward plus the policy forward and per-step bookkeeping;
        the 7.5M and 300M policies are randomly initialized.
    }
    \label{tab:step_time_full}
\end{table}

%% file: sec_app/c_experimental_details.tex
\section{Experimental Details}
\label{app:experimental_details}

\subsection{Datasets and Prompt Templates}
\label{app:datasets}

\noindent\textbf{Datasets.}
We train on the GSM8K training split (\texttt{openai/gsm8k}) and
the MATH training split (\texttt{ankner/math-500}).
For MATH, we exclude 15 examples whose gold answers cannot be parsed
by the answer verifier, leaving 7,485 examples.
The resulting 14,958 training examples are concatenated and shuffled
with seed 42.
During training, prompts exceeding 256 tokens are left-truncated
to retain the last 256 tokens.
Evaluation prompts are not truncated, and no evaluation examples
are used for training.

\Cref{tab:dataset_details} summarizes the dataset sizes.
All evaluations use zero-shot prompting unless otherwise stated.
GSM8K uses the \texttt{openai/gsm8k} test split, MATH-500 uses
\texttt{HuggingFaceH4/MATH-500}, HumanEval uses
\texttt{openai/openai\_humaneval}, and MBPP uses the sanitized
test split of \texttt{google-research-datasets/mbpp}.

\begin{table}[t]
    \centering
    \small
    \begin{tabularx}{\columnwidth}{X rr}
        \toprule
        Dataset & Training & Evaluation \\
        \midrule
        GSM8K & 7,473 & 1,319 \\
        MATH & 7,485 & -- \\
        MATH-500 & -- & 500 \\
        HumanEval & -- & 164 \\
        MBPP & -- & 257 \\
        \midrule
        Total & 14,958 & 2,240 \\
        \bottomrule
    \end{tabularx}
    \caption{Training and evaluation dataset sizes.}
    \label{tab:dataset_details}
\end{table}

\noindent\textbf{Prompt templates.}
We use task-specific instruction prompts with the backbone's chat template.
For GSM8K and MATH, we use the same prompt for training and evaluation,
instructing the model to solve problems step by step and provide the
final answer in \verb|\boxed{}|.
For HumanEval and MBPP, we follow the prompts from
\texttt{lm-eval}.
HumanEval provides a function signature and docstring, while MBPP
provides a task description and test cases.
For both tasks, generation continues from a partially provided
code block.

\noindent\textbf{Answer verification.}
For GSM8K and MATH, we extract and verify the final boxed answer
using \texttt{math-verify} v0.8.0.
The same procedure is used to select correct reference trajectories
for training.

\subsection{Policy Optimization}
\label{app:optimization}

We train the full policy jointly with AdamW.
Training takes approximately 3 hours and 15 minutes on a single
NVIDIA H200 GPU.
Unless otherwise specified, all policy-training experiments use the
hyperparameters in \Cref{tab:optimization_details}.

\begin{table}[t]
    \centering
    \small
    \begin{tabularx}{\columnwidth}{X r}
        \toprule
        Hyperparameter & Value \\
        \midrule
        Optimizer & AdamW \\
        Learning rate & $3\times10^{-5}$ \\
        Adam $(\beta_1, \beta_2)$ & $(0.9, 0.99)$ \\
        Adam $\epsilon$ & $10^{-8}$ \\
        Weight decay & $0.1$ \\
        LR scheduler & Cosine decay \\
        Warmup & 100 optimizer steps \\
        Gradient clipping & $0.2$ max norm \\
        Batch size & 16 / GPU \\
        Gradient accumulation & 1 \\
        Epochs & 1 \\
        Dropout & 0 \\
        Seed & 123 \\
        \bottomrule
    \end{tabularx}
    \caption{Optimization hyperparameters used for policy training.}
    \label{tab:optimization_details}
\end{table}

\subsection{Training and Decoding Configuration}
\label{app:training_configuration}

Following \citet{jazbec2026learning}, we train the policy on the training
splits of GSM8K and MATH.
We retain only reference trajectories whose final answer is correct,
yielding 9,169 trajectories.
No code-generation examples are used for policy training; HumanEval and MBPP
therefore evaluate out-of-domain generalization from mathematical reasoning
to code.

Reference trajectories are generated by Fast-dLLM
\citep{wu2026fastdllm} with a confidence threshold of $0.9$.
We use generation length $L=256$ and block length $32$ for both trajectory
collection and the main evaluation.
For the length-transfer evaluation at $L=512$ (\Cref{sec:length_generalization}),
we additionally apply End-of-Text Prediction (EoTP) \citep{bao2026learning}
to all methods: once an end-of-text token has been committed, all remaining
masked positions after it are filled with end-of-text tokens, and decoding
terminates.
As a result, the NFE at $L=512$ can be smaller than $L$ even for one-by-one decoding.
The trajectory-generation temperature is set to zero.

The frozen LLaDA-8B-Instruct backbone is loaded in bfloat16, while policy
computation is performed in float32.
The backbone parameters remain frozen throughout policy training.

\begin{table}[t]
    \centering
    \small
    \begin{tabularx}{\columnwidth}{X r}
        \toprule
        Setting & Value \\
        \midrule
        Training data & GSM8K + MATH train \\
        Reference decoder & Fast-dLLM \\
        Reference threshold & $0.9$ \\
        Generation length & 256 \\
        Block length & 32 \\
        Training temperature & 0 \\
        Backbone precision & bfloat16 \\
        Policy precision & float32 \\
        \bottomrule
    \end{tabularx}
    \caption{Training-trajectory and numerical configuration.}
    \label{tab:training_configuration}
\end{table}

\subsection{Parameter Grids}
\label{app:param_grids}

\Cref{tab:param_grids} summarizes the commitment rule of each method and
the parameter swept to obtain operating points.
OnTrack (Ours), Learn2PD, and Fast-dLLM commit every eligible position whose score
exceeds a threshold $\gamma$, which we sweep over the same grid
$\gamma\in\{0.5, 0.6, 0.7, 0.8, 0.85, 0.9, 0.94, 0.96, 0.98, 0.99\}$
for all three methods.
EB-Sampler \citep{ben-hamu2025accelerated} sorts positions by confidence
and commits the longest prefix whose total entropy minus its maximum entropy
does not exceed a budget $\epsilon$, which we sweep over
$\epsilon\in\{0, 0.001, 0.003, 0.01, 0.03, 0.1, 0.3, 1, 3, 10\}$.

\begin{table}[h]
    \centering
    \small
    \begin{tabularx}{\columnwidth}{l X c}
        \toprule
        Method & Score / commitment rule & Param. \\
        \midrule
        OnTrack (Ours)       & learned policy probability $\sigma(\ell_{t,i})>\gamma$ & $\gamma$ \\
        Learn2PD   & learned filter probability $>\gamma$                  & $\gamma$ \\
        Fast-dLLM  & top-1 token probability $>\gamma$                     & $\gamma$ \\
        EB-Sampler & entropy-bounded prefix of confidence-sorted positions  & $\epsilon$ \\
        \bottomrule
    \end{tabularx}
    \caption{Commitment rules and swept parameters.}
    \label{tab:param_grids}
\end{table}

\begin{figure*}[t]
    \centering
    \includegraphics[width=\textwidth]{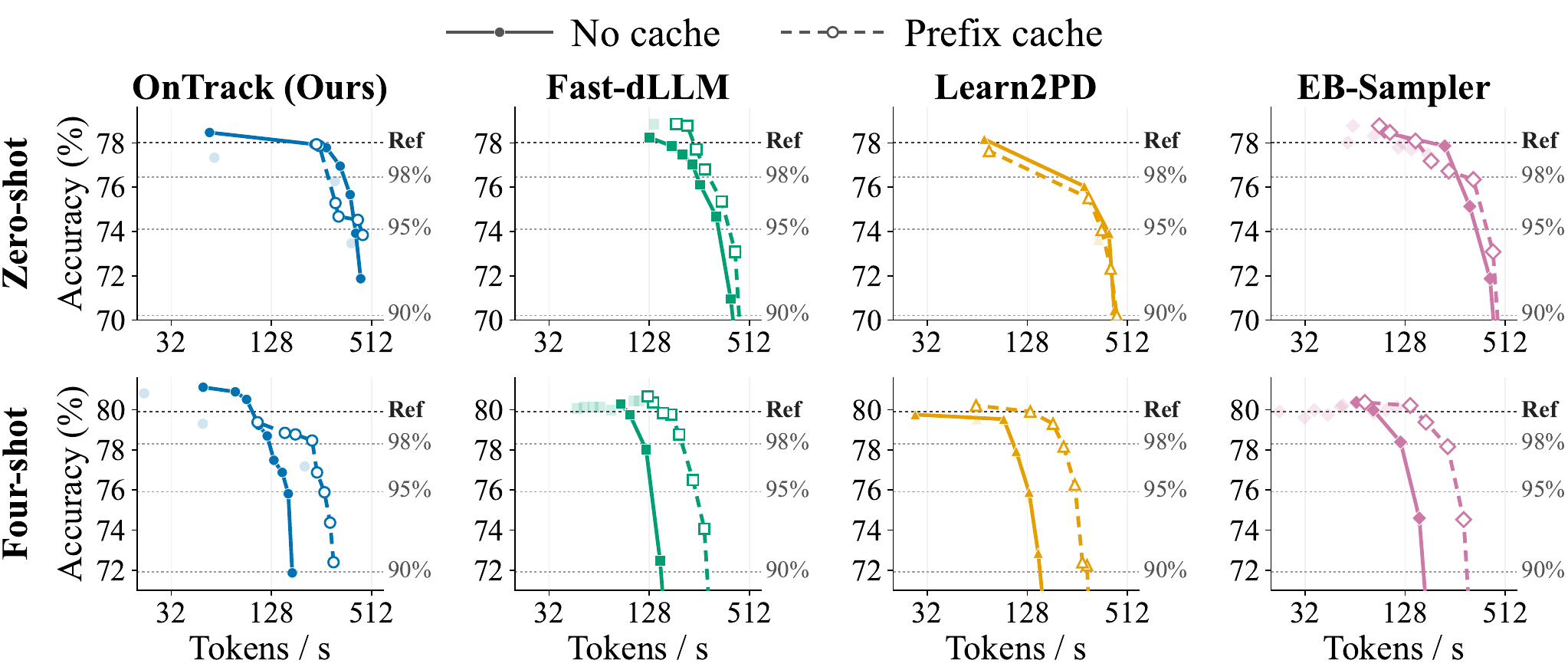}
    \caption{
        Accuracy versus throughput on GSM8K with zero-shot (top) and
        four-shot (bottom) prompting.
        Solid and dashed lines indicate decoding without and with
        prefix caching, respectively.
        Horizontal lines mark the uncached one-by-one reference accuracy
        and its 98\%, 95\%, and 90\% retention thresholds.
    }
    \label{fig:prefix_cache_accuracy}
\end{figure*}

\begin{figure}[t]
    \centering
    \includegraphics[width=\linewidth]{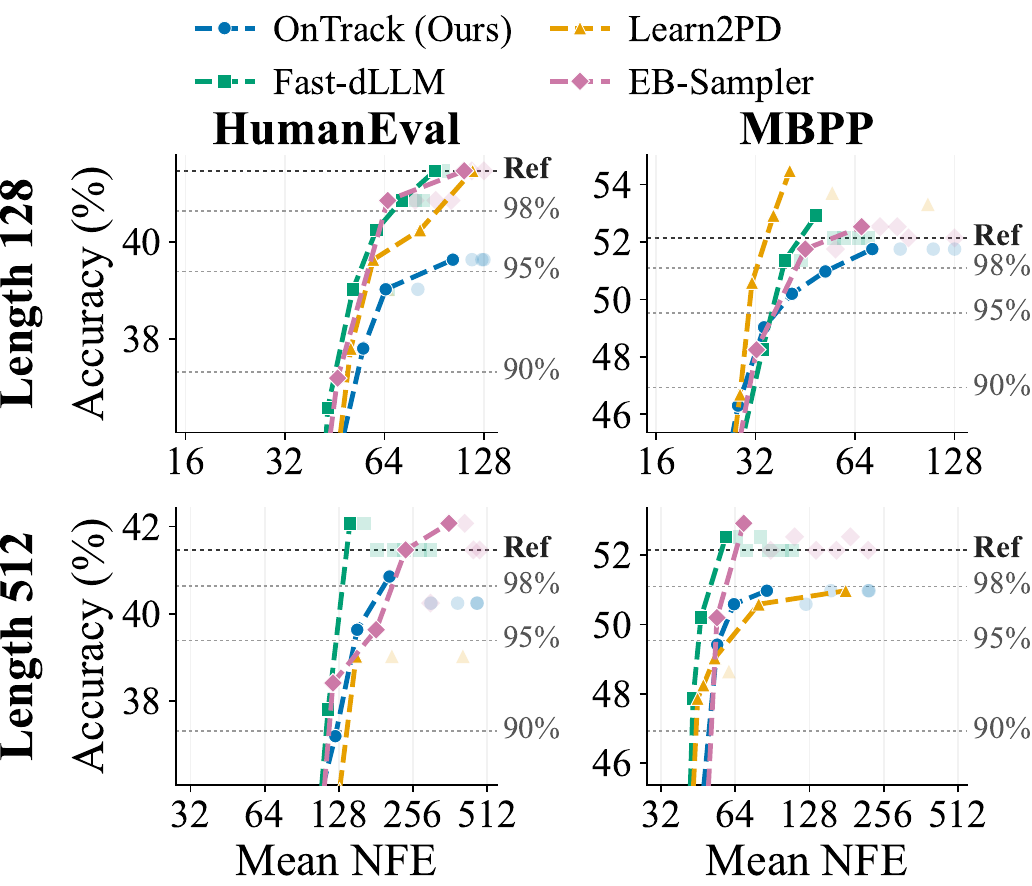}
    \caption{
        Accuracy versus NFE on HumanEval and MBPP at generation lengths
        $L=128$ (top) and $L=512$ (bottom) for a policy trained at $L=256$.
    }
    \label{fig:length_results_code}
\end{figure}

\subsection{Throughput Measurement}
\label{app:throughput}

All throughput measurements are performed on a single NVIDIA H200 GPU
(141\,GB) with batch size 1, using LLaDA-8B-Instruct in bfloat16 with
FlashAttention-2 \citep{dao2024flashattention}.
TPS is computed as the number of generated tokens, excluding tokenizer
special tokens, divided by the sampler wall-clock time, which includes
both backbone and policy evaluations but excludes tokenization,
detokenization, and answer scoring.
For each prompt, we call \texttt{torch.cuda.synchronize} immediately
before and after the sampler call and time the interval with a
high-resolution host timer.
No warm-up runs are performed; each prompt is decoded once (a single
pass over the full evaluation set), with only one generation process
per GPU.
Timing is aggregated per dataset and decoding configuration as the
total number of generated tokens over the total sampler time, i.e., a
token-weighted average rather than a mean of per-prompt rates.

\subsection{Additional Results}
\label{app:length_results}

For the prefix caching experiments (\Cref{sec:prefix_cache}),
\Cref{fig:prefix_cache_accuracy} shows the accuracy--throughput frontiers
on GSM8K under zero-shot and four-shot prompting.
Our policy exhibits greater accuracy sensitivity to caching than the
baselines, while four-shot prompting yields substantial throughput
improvements with caching.

For generalization across sequence lengths (\Cref{sec:length_generalization}),
\Cref{fig:length_results_code} shows the frontiers on HumanEval and MBPP at
$L=128$ and $L=512$, complementing \Cref{fig:length_results}.
The corresponding operating points, together with those for all other
figures, are listed in \Cref{app:numerical_results}.

%% file: sec_app/d_numerical_results.tex
\section{Numerical Results}
\label{app:numerical_results}

This section lists all measured operating points underlying the figures.
\Cref{tab:results_supervision,tab:results_supervision_extra} correspond to
\Cref{fig:teacher_signal,fig:teacher_signal_extra};
\Cref{tab:results_input,tab:results_input_extra} to
\Cref{fig:input_ablation,fig:input_ablation_extra};
\Cref{tab:results_main_math,tab:results_main_code} to
\Cref{fig:main_results};
\Cref{tab:prefix_cache_gsm8k_zero_shot_full,tab:prefix_cache_gsm8k_four_shot_full} to
\Cref{fig:prefix_cache_accuracy}; 
\Cref{tab:results_dream_math,tab:results_dream_code} to
\Cref{fig:dream_results}; 
and
\Cref{tab:results_len128_math,tab:results_len128_code,tab:results_len512_math,tab:results_len512_code}
to \Cref{fig:length_results,fig:length_results_code}.
\begin{table*}[t]
    \centering
    \small
    \input{tab/teacher_signal_dim128_gsm8k_mbpp}
        \caption{All operating points for the supervision study on GSM8K and
        MBPP (\Cref{fig:teacher_signal}).}
    \label{tab:results_supervision}
\end{table*}

\begin{table*}[t]
    \centering
    \small
    \input{tab/input_ablation_gsm8k_mbpp}
    \caption{All operating points for the policy-input study on GSM8K and
    MBPP (\Cref{fig:input_ablation}).}
    \label{tab:results_input}
\end{table*}

\begin{table*}[t]
    \centering
    \small
    \input{tab/teacher_signal_dim128_math_humaneval}
    \caption{All operating points for the supervision study on MATH-500
        and HumanEval (\Cref{fig:teacher_signal_extra}).}
    \label{tab:results_supervision_extra}
\end{table*}

\begin{table*}[t]
    \centering
    \small
    \input{tab/input_ablation_math_humaneval}
    \caption{All operating points for the policy-input study on MATH-500
        and HumanEval (\Cref{fig:input_ablation_extra}).}
    \label{tab:results_input_extra}
\end{table*}

\begin{table*}[t]
    \centering
    \small
    \input{tab/llada8b_gsm8k_math}
    \caption{All operating points on GSM8K and MATH-500 with
        LLaDA-8B-Instruct (\Cref{fig:main_results}).
        }
    \label{tab:results_main_math}
\end{table*}

\begin{table*}[t]
    \centering
    \small
    \input{tab/llada8b_humaneval_mbpp}
    \caption{All operating points on HumanEval and MBPP with
        LLaDA-8B-Instruct (\Cref{fig:main_results}).}
    \label{tab:results_main_code}
\end{table*}

\begin{table*}[t]
    \centering
    \small
    \input{tab/prefix_cache_gsm8k_zero_shot_full}
    \caption{Full GSM8K results with and without prefix caching under zero-shot prompting (\Cref{tab:prefix_cache}).}
    \label{tab:prefix_cache_gsm8k_zero_shot_full}
\end{table*}

\begin{table*}[t]
    \centering
    \small
    \input{tab/prefix_cache_gsm8k_four_shot_full}
    \caption{Full GSM8K results with and without prefix caching under four-shot prompting (\Cref{tab:prefix_cache}).}
    \label{tab:prefix_cache_gsm8k_four_shot_full}
\end{table*}

\begin{table*}[t]
    \centering
    \small
    \input{tab/dream7b_gsm8k_math}
    \caption{All operating points on GSM8K and MATH-500 with
        Dream-7B-Instruct (\Cref{fig:dream_results}).}
    \label{tab:results_dream_math}
\end{table*}

\begin{table*}[t]
    \centering
    \small
    \input{tab/dream7b_humaneval_mbpp}
    \caption{All operating points on HumanEval and MBPP with
        Dream-7B-Instruct (\Cref{fig:dream_results}).}
    \label{tab:results_dream_code}
\end{table*}

\begin{table*}[t]
    \centering
    \small
    \input{tab/llada8b_length128_gsm8k_math}
    \caption{All operating points on GSM8K and MATH-500 at $L=128$
        (\Cref{fig:length_results}).}
    \label{tab:results_len128_math}
\end{table*}

\begin{table*}[t]
    \centering
    \small
    \input{tab/llada8b_length128_humaneval_mbpp}
    \caption{All operating points on HumanEval and MBPP at $L=128$
        (\Cref{fig:length_results_code}).}
    \label{tab:results_len128_code}
\end{table*}

\begin{table*}[t]
    \centering
    \small
    \input{tab/llada8b_length512_gsm8k_math}
    \caption{All operating points on GSM8K and MATH-500 at $L=512$
        (\Cref{fig:length_results}).}
    \label{tab:results_len512_math}
\end{table*}

\begin{table*}[t]
    \centering
    \small
    \input{tab/llada8b_length512_humaneval_mbpp}
    \caption{All operating points on HumanEval and MBPP at $L=512$
        (\Cref{fig:length_results_code}).}
    \label{tab:results_len512_code}
\end{table*}

\begin{table*}[t]
    \centering
    \small
    \input{tab/llada8b_ci_summary}
       \caption{NFE speedup of OnTrack over each LLaDA-8B baseline, defined as the
        baseline's NFE divided by OnTrack's NFE to reach 90\%, 95\%, or 98\% of the
        one-by-one highest-confidence reference accuracy (GSM8K 78.0, MATH-500 38.8,
        HumanEval 45.1, MBPP 52.9; the same target for all methods). NFE-to-target is
        the smallest measured mean NFE whose accuracy meets the target, on the common
        threshold grid without interpolation. Brackets give paired bootstrap 95\%
        confidence intervals over evaluation problems (10{,}000 resamples); bold marks
        intervals that exclude 1. n.r.: OnTrack does not reach the target; --: neither
        method does.}
    \label{tab:results_llada_ci}
\end{table*}

%% file: tab/teacher_signal_dim128_gsm8k_mbpp.tex
\begin{tabular}{ll rr @{\hspace{1.5em}} ll rr}
    \toprule
    \multicolumn{4}{c}{GSM8K} & \multicolumn{4}{c}{MBPP} \\
    \cmidrule(lr){1-4}\cmidrule(lr){5-8}
    Method & Param. & Acc. & NFE & Method & Param. & Acc. & NFE \\
    \midrule
    Teacher + terminal & 0.99 & 79.2 & 256.0 & Teacher + terminal & 0.99 & 52.5 & 256.0 \\
     & 0.98 & 78.8 & 100.1 &  & 0.98 & 52.5 & 187.7 \\
     & 0.96 & 78.4 & 66.0 &  & 0.96 & 52.5 & 127.6 \\
     & 0.94 & 77.8 & 55.9 &  & 0.94 & 52.5 & 100.5 \\
     & 0.9 & 75.1 & 46.8 &  & 0.9 & 51.4 & 70.4 \\
     & 0.85 & 74.0 & 40.6 &  & 0.85 & 49.0 & 56.6 \\
     & 0.8 & 73.0 & 37.1 &  & 0.8 & 46.3 & 48.0 \\
     & 0.7 & 67.8 & 33.0 &  & 0.7 & 33.5 & 39.0 \\
     & 0.6 & 63.2 & 30.5 &  & 0.6 & 26.5 & 30.0 \\
     & 0.5 & 55.4 & 28.1 &  & 0.5 & 18.3 & 24.8 \\
    \midrule
    Teacher + contraction & 0.99 & 78.1 & 256.0 & Teacher + contraction & 0.99 & 53.7 & 256.0 \\
     & 0.98 & 78.2 & 105.0 &  & 0.98 & 53.7 & 179.3 \\
     & 0.96 & 77.9 & 69.7 &  & 0.96 & 53.7 & 117.6 \\
     & 0.94 & 78.2 & 60.2 &  & 0.94 & 53.7 & 90.1 \\
     & 0.9 & 77.8 & 51.2 &  & 0.9 & 53.7 & 64.8 \\
     & 0.85 & 77.3 & 45.3 &  & 0.85 & 51.4 & 50.7 \\
     & 0.8 & 76.6 & 41.7 &  & 0.8 & 47.5 & 44.4 \\
     & 0.7 & 75.1 & 36.8 &  & 0.7 & 44.0 & 39.5 \\
     & 0.6 & 74.2 & 33.6 &  & 0.6 & 36.2 & 36.9 \\
     & 0.5 & 71.7 & 31.3 &  & 0.5 & 30.4 & 34.2 \\
    \midrule
    Reference + terminal & 0.99 & 78.6 & 256.0 & Reference + terminal & 0.99 & 51.8 & 256.0 \\
     & 0.98 & 78.8 & 70.8 &  & 0.98 & 51.8 & 192.4 \\
     & 0.96 & 76.9 & 51.8 &  & 0.96 & 51.8 & 116.0 \\
     & 0.94 & 76.2 & 45.7 &  & 0.94 & 51.0 & 81.3 \\
     & 0.9 & 74.7 & 39.8 &  & 0.9 & 48.2 & 54.1 \\
     & 0.85 & 73.6 & 35.7 &  & 0.85 & 43.2 & 42.7 \\
     & 0.8 & 70.6 & 33.6 &  & 0.8 & 36.2 & 38.5 \\
     & 0.7 & 67.2 & 30.5 &  & 0.7 & 29.2 & 32.2 \\
     & 0.6 & 59.9 & 28.5 &  & 0.6 & 21.8 & 27.8 \\
     & 0.5 & 54.0 & 26.4 &  & 0.5 & 17.1 & 20.3 \\
    \midrule
    Reference + contraction & 0.99 & 78.5 & 256.0 & Reference + contraction & 0.99 & 53.7 & 256.0 \\
     & 0.98 & 78.4 & 109.0 &  & 0.98 & 53.7 & 196.9 \\
     & 0.96 & 78.8 & 70.6 &  & 0.96 & 53.7 & 131.9 \\
     & 0.94 & 77.9 & 60.6 &  & 0.94 & 53.7 & 103.3 \\
     & 0.9 & 77.8 & 51.7 &  & 0.9 & 54.1 & 73.4 \\
     & 0.85 & 76.3 & 45.8 &  & 0.85 & 54.1 & 57.1 \\
     & 0.8 & 77.0 & 42.1 &  & 0.8 & 52.5 & 48.6 \\
     & 0.7 & 75.7 & 37.1 &  & 0.7 & 45.9 & 42.4 \\
     & 0.6 & 73.9 & 34.2 &  & 0.6 & 42.0 & 39.4 \\
     & 0.5 & 71.9 & 31.8 &  & 0.5 & 37.0 & 35.9 \\
    \bottomrule
\end{tabular}

%% file: tab/input_ablation_gsm8k_mbpp.tex
\begin{tabular}{ll rr @{\hspace{1.5em}} ll rr}
    \toprule
    \multicolumn{4}{c}{GSM8K} & \multicolumn{4}{c}{MBPP} \\
    \cmidrule(lr){1-4}\cmidrule(lr){5-8}
    Method & Param. & Acc. & NFE & Method & Param. & Acc. & NFE \\
    \midrule
    Conf. & 0.99 & 78.1 & 256.0 & Conf. & 0.99 & 53.3 & 256.0 \\
     & 0.98 & 78.2 & 158.4 &  & 0.98 & 53.3 & 162.2 \\
     & 0.96 & 78.1 & 72.5 &  & 0.96 & 52.5 & 65.8 \\
     & 0.94 & 77.9 & 62.0 &  & 0.94 & 52.5 & 58.8 \\
     & 0.9 & 77.0 & 52.9 &  & 0.9 & 51.4 & 51.9 \\
     & 0.85 & 75.3 & 47.2 &  & 0.85 & 50.2 & 47.3 \\
     & 0.8 & 75.1 & 43.6 &  & 0.8 & 49.4 & 44.2 \\
     & 0.7 & 75.1 & 39.0 &  & 0.7 & 44.0 & 41.0 \\
     & 0.6 & 72.4 & 36.2 &  & 0.6 & 41.6 & 42.0 \\
     & 0.5 & 72.6 & 33.5 &  & 0.5 & 36.6 & 41.3 \\
    \midrule
    Hidden & 0.99 & 77.5 & 256.0 & Hidden & 0.99 & 51.0 & 256.0 \\
     & 0.98 & 77.4 & 227.6 &  & 0.98 & 51.0 & 197.8 \\
     & 0.96 & 77.6 & 138.6 &  & 0.96 & 51.0 & 143.8 \\
     & 0.94 & 77.6 & 105.3 &  & 0.94 & 50.6 & 124.0 \\
     & 0.9 & 76.4 & 77.1 &  & 0.9 & 50.6 & 99.4 \\
     & 0.85 & 74.5 & 61.3 &  & 0.85 & 49.8 & 83.6 \\
     & 0.8 & 75.4 & 51.9 &  & 0.8 & 48.2 & 73.1 \\
     & 0.7 & 69.8 & 42.2 &  & 0.7 & 42.0 & 59.0 \\
     & 0.6 & 67.5 & 37.6 &  & 0.6 & 37.7 & 51.0 \\
     & 0.5 & 64.0 & 33.8 &  & 0.5 & 31.1 & 41.4 \\
    \midrule
    Conf.+Margin & 0.99 & 77.9 & 255.9 & Conf.+Margin & 0.99 & 52.9 & 255.9 \\
     & 0.98 & 77.9 & 99.2 &  & 0.98 & 52.9 & 95.3 \\
     & 0.96 & 77.1 & 66.2 &  & 0.96 & 51.4 & 60.9 \\
     & 0.94 & 77.4 & 58.3 &  & 0.94 & 51.4 & 54.9 \\
     & 0.9 & 76.0 & 50.7 &  & 0.9 & 49.8 & 49.1 \\
     & 0.85 & 75.7 & 45.5 &  & 0.85 & 48.6 & 44.1 \\
     & 0.8 & 75.5 & 42.0 &  & 0.8 & 48.2 & 43.6 \\
     & 0.7 & 74.5 & 37.6 &  & 0.7 & 43.6 & 41.5 \\
     & 0.6 & 73.0 & 34.3 &  & 0.6 & 43.2 & 40.1 \\
     & 0.5 & 70.8 & 32.0 &  & 0.5 & 40.1 & 38.7 \\
    \midrule
    Conf.+Hidden & 0.99 & 78.5 & 256.0 & Conf.+Hidden & 0.99 & 53.7 & 256.0 \\
     & 0.98 & 78.4 & 109.0 &  & 0.98 & 53.7 & 196.9 \\
     & 0.96 & 78.8 & 70.6 &  & 0.96 & 53.7 & 131.9 \\
     & 0.94 & 77.9 & 60.6 &  & 0.94 & 53.7 & 103.3 \\
     & 0.9 & 77.8 & 51.7 &  & 0.9 & 54.1 & 73.4 \\
     & 0.85 & 76.3 & 45.8 &  & 0.85 & 54.1 & 57.1 \\
     & 0.8 & 77.0 & 42.1 &  & 0.8 & 52.5 & 48.6 \\
     & 0.7 & 75.7 & 37.1 &  & 0.7 & 45.9 & 42.4 \\
     & 0.6 & 73.9 & 34.2 &  & 0.6 & 42.0 & 39.4 \\
     & 0.5 & 71.9 & 31.8 &  & 0.5 & 37.0 & 35.9 \\
    \bottomrule
\end{tabular}

%% file: tab/teacher_signal_dim128_math_humaneval.tex
\begin{tabular}{ll rr @{\hspace{1.5em}} ll rr}
    \toprule
    \multicolumn{4}{c}{MATH-500} & \multicolumn{4}{c}{HumanEval} \\
    \cmidrule(lr){1-4}\cmidrule(lr){5-8}
    Method & Param. & Acc. & NFE & Method & Param. & Acc. & NFE \\
    \midrule
    Teacher + terminal & 0.99 & 38.0 & 256.0 & Teacher + terminal & 0.99 & 42.1 & 256.0 \\
     & 0.98 & 38.2 & 137.6 &  & 0.98 & 42.1 & 246.0 \\
     & 0.96 & 36.6 & 93.9 &  & 0.96 & 42.1 & 213.3 \\
     & 0.94 & 37.0 & 81.3 &  & 0.94 & 41.5 & 175.4 \\
     & 0.9 & 36.0 & 68.6 &  & 0.9 & 39.6 & 126.3 \\
     & 0.85 & 36.8 & 60.2 &  & 0.85 & 36.6 & 94.7 \\
     & 0.8 & 34.0 & 54.9 &  & 0.8 & 34.1 & 76.5 \\
     & 0.7 & 31.2 & 47.6 &  & 0.7 & 23.8 & 53.7 \\
     & 0.6 & 26.8 & 42.9 &  & 0.6 & 14.6 & 40.7 \\
     & 0.5 & 22.8 & 38.1 &  & 0.5 & 9.1 & 29.1 \\
    \midrule
    Teacher + contraction & 0.99 & 38.4 & 256.0 & Teacher + contraction & 0.99 & 42.7 & 256.0 \\
     & 0.98 & 38.2 & 128.2 &  & 0.98 & 42.7 & 241.8 \\
     & 0.96 & 38.4 & 88.9 &  & 0.96 & 42.7 & 192.6 \\
     & 0.94 & 38.6 & 78.5 &  & 0.94 & 42.7 & 153.8 \\
     & 0.9 & 37.8 & 67.9 &  & 0.9 & 42.7 & 112.7 \\
     & 0.85 & 37.0 & 59.6 &  & 0.85 & 37.2 & 86.4 \\
     & 0.8 & 36.4 & 55.5 &  & 0.8 & 34.8 & 74.4 \\
     & 0.7 & 32.4 & 49.1 &  & 0.7 & 31.7 & 58.6 \\
     & 0.6 & 31.6 & 44.3 &  & 0.6 & 29.3 & 48.8 \\
     & 0.5 & 31.6 & 41.2 &  & 0.5 & 22.0 & 40.0 \\
    \midrule
    Reference + terminal & 0.99 & 39.8 & 256.0 & Reference + terminal & 0.99 & 39.0 & 256.0 \\
     & 0.98 & 39.4 & 88.1 &  & 0.98 & 39.0 & 244.4 \\
     & 0.96 & 39.2 & 68.4 &  & 0.96 & 39.0 & 186.0 \\
     & 0.94 & 35.8 & 62.0 &  & 0.94 & 37.8 & 140.8 \\
     & 0.9 & 35.6 & 54.5 &  & 0.9 & 36.6 & 91.3 \\
     & 0.85 & 32.8 & 49.3 &  & 0.85 & 31.7 & 67.1 \\
     & 0.8 & 31.6 & 46.6 &  & 0.8 & 21.3 & 54.2 \\
     & 0.7 & 32.6 & 41.9 &  & 0.7 & 18.3 & 41.0 \\
     & 0.6 & 28.6 & 38.9 &  & 0.6 & 12.2 & 31.3 \\
     & 0.5 & 23.4 & 34.9 &  & 0.5 & 8.5 & 22.4 \\
    \midrule
    Reference + contraction & 0.99 & 37.2 & 256.0 & Reference + contraction & 0.99 & 43.3 & 256.0 \\
     & 0.98 & 37.2 & 131.2 &  & 0.98 & 43.3 & 247.1 \\
     & 0.96 & 37.6 & 88.7 &  & 0.96 & 43.3 & 212.9 \\
     & 0.94 & 37.8 & 77.8 &  & 0.94 & 43.9 & 176.5 \\
     & 0.9 & 38.0 & 68.0 &  & 0.9 & 43.3 & 130.0 \\
     & 0.85 & 36.8 & 60.5 &  & 0.85 & 42.1 & 99.1 \\
     & 0.8 & 37.0 & 55.8 &  & 0.8 & 41.5 & 82.9 \\
     & 0.7 & 33.8 & 49.6 &  & 0.7 & 35.4 & 65.6 \\
     & 0.6 & 33.8 & 45.0 &  & 0.6 & 33.5 & 54.8 \\
     & 0.5 & 34.2 & 41.7 &  & 0.5 & 25.6 & 44.0 \\
    \bottomrule
\end{tabular}

%% file: tab/input_ablation_math_humaneval.tex
\begin{tabular}{ll rr @{\hspace{1.5em}} ll rr}
    \toprule
    \multicolumn{4}{c}{MATH-500} & \multicolumn{4}{c}{HumanEval} \\
    \cmidrule(lr){1-4}\cmidrule(lr){5-8}
    Method & Param. & Acc. & NFE & Method & Param. & Acc. & NFE \\
    \midrule
    Confidence & 0.99 & 37.6 & 256.0 & Confidence & 0.99 & 42.7 & 256.0 \\
     & 0.98 & 37.8 & 170.0 &  & 0.98 & 42.7 & 182.9 \\
     & 0.96 & 38.2 & 98.4 &  & 0.96 & 42.7 & 111.9 \\
     & 0.94 & 37.6 & 86.5 &  & 0.94 & 40.9 & 100.1 \\
     & 0.9 & 38.2 & 76.7 &  & 0.9 & 42.1 & 86.5 \\
     & 0.85 & 37.2 & 69.5 &  & 0.85 & 40.9 & 78.5 \\
     & 0.8 & 37.0 & 65.2 &  & 0.8 & 34.8 & 73.1 \\
     & 0.7 & 36.4 & 58.9 &  & 0.7 & 33.5 & 65.1 \\
     & 0.6 & 35.8 & 54.8 &  & 0.6 & 26.8 & 59.0 \\
     & 0.5 & 31.6 & 50.9 &  & 0.5 & 25.6 & 53.6 \\
    \midrule
    Hidden & 0.99 & 36.2 & 256.0 & Hidden & 0.99 & 37.8 & 256.0 \\
     & 0.98 & 36.2 & 240.2 &  & 0.98 & 37.8 & 247.7 \\
     & 0.96 & 35.6 & 162.6 &  & 0.96 & 37.8 & 221.3 \\
     & 0.94 & 35.6 & 129.6 &  & 0.94 & 37.8 & 201.8 \\
     & 0.9 & 36.8 & 101.2 &  & 0.9 & 37.8 & 168.6 \\
     & 0.85 & 34.6 & 82.6 &  & 0.85 & 37.2 & 141.5 \\
     & 0.8 & 32.2 & 70.5 &  & 0.8 & 34.8 & 122.3 \\
     & 0.7 & 31.8 & 58.5 &  & 0.7 & 32.9 & 99.2 \\
     & 0.6 & 28.0 & 49.0 &  & 0.6 & 29.9 & 78.6 \\
     & 0.5 & 27.2 & 42.6 &  & 0.5 & 25.6 & 64.8 \\
    \midrule
    Confidence + Margin & 0.99 & 40.0 & 255.7 & Confidence + Margin & 0.99 & 43.9 & 255.8 \\
     & 0.98 & 39.8 & 122.3 &  & 0.98 & 43.3 & 133.4 \\
     & 0.96 & 39.8 & 90.6 &  & 0.96 & 42.7 & 103.7 \\
     & 0.94 & 39.0 & 81.3 &  & 0.94 & 41.5 & 93.3 \\
     & 0.9 & 38.4 & 72.8 &  & 0.9 & 42.1 & 82.2 \\
     & 0.85 & 35.6 & 66.1 &  & 0.85 & 40.2 & 74.1 \\
     & 0.8 & 36.0 & 61.4 &  & 0.8 & 36.6 & 69.7 \\
     & 0.7 & 36.0 & 55.4 &  & 0.7 & 32.9 & 60.2 \\
     & 0.6 & 33.0 & 50.6 &  & 0.6 & 29.9 & 55.0 \\
     & 0.5 & 32.8 & 47.8 &  & 0.5 & 23.8 & 50.3 \\
    \midrule
    Confidence + Hidden & 0.99 & 37.2 & 256.0 & Confidence + Hidden & 0.99 & 43.3 & 256.0 \\
     & 0.98 & 37.2 & 131.2 &  & 0.98 & 43.3 & 247.1 \\
     & 0.96 & 37.6 & 88.7 &  & 0.96 & 43.3 & 212.9 \\
     & 0.94 & 37.8 & 77.8 &  & 0.94 & 43.9 & 176.5 \\
     & 0.9 & 38.0 & 68.0 &  & 0.9 & 43.3 & 130.0 \\
     & 0.85 & 36.8 & 60.5 &  & 0.85 & 42.1 & 99.1 \\
     & 0.8 & 37.0 & 55.8 &  & 0.8 & 41.5 & 82.9 \\
     & 0.7 & 33.8 & 49.6 &  & 0.7 & 35.4 & 65.6 \\
     & 0.6 & 33.8 & 45.0 &  & 0.6 & 33.5 & 54.8 \\
     & 0.5 & 34.2 & 41.7 &  & 0.5 & 25.6 & 44.0 \\
    \bottomrule
\end{tabular}

%% file: tab/llada8b_gsm8k_math.tex
\begin{tabular}{ll rrr @{\hspace{1.5em}} ll rrr}
    \toprule
    \multicolumn{5}{c}{GSM8K} & \multicolumn{5}{c}{MATH-500} \\
    \cmidrule(lr){1-5}\cmidrule(lr){6-10}
    Method & Param. & Acc. & NFE & TPS & Method & Param. & Acc. & NFE & TPS \\
    \midrule
    OnTrack (Ours) & 0.99 & 78.5 & 256.0 & 54 & OnTrack (Ours) & 0.99 & 37.2 & 256.0 & 53 \\
     & 0.98 & 78.4 & 109.0 & 127 &  & 0.98 & 37.2 & 131.2 & 103 \\
     & 0.96 & 78.8 & 70.6 & 196 &  & 0.96 & 37.6 & 88.7 & 153 \\
     & 0.94 & 77.9 & 60.6 & 230 &  & 0.94 & 37.8 & 77.8 & 177 \\
     & 0.9 & 77.8 & 51.7 & 274 &  & 0.9 & 38.0 & 68.0 & 200 \\
     & 0.85 & 76.3 & 45.8 & 308 &  & 0.85 & 36.8 & 60.5 & 225 \\
     & 0.8 & 77.0 & 42.1 & 331 &  & 0.8 & 37.0 & 55.8 & 244 \\
     & 0.7 & 75.7 & 37.1 & 381 &  & 0.7 & 33.8 & 49.6 & 274 \\
     & 0.6 & 73.9 & 34.2 & 411 &  & 0.6 & 33.8 & 45.0 & 303 \\
     & 0.5 & 71.9 & 31.8 & 440 &  & 0.5 & 34.2 & 41.7 & 327 \\
    \midrule
    Fast-dLLM & 0.99 & 78.2 & 118.1 & 128 & Fast-dLLM & 0.99 & 38.8 & 150.2 & 99 \\
     & 0.98 & 78.3 & 107.9 & 140 &  & 0.98 & 38.8 & 138.8 & 106 \\
     & 0.96 & 78.2 & 94.5 & 160 &  & 0.96 & 38.8 & 124.2 & 119 \\
     & 0.94 & 77.9 & 86.3 & 174 &  & 0.94 & 39.0 & 115.3 & 127 \\
     & 0.9 & 77.5 & 74.4 & 202 &  & 0.9 & 38.4 & 102.6 & 143 \\
     & 0.85 & 77.0 & 65.6 & 232 &  & 0.85 & 38.8 & 92.7 & 157 \\
     & 0.8 & 76.1 & 58.2 & 258 &  & 0.8 & 38.0 & 83.7 & 175 \\
     & 0.7 & 74.7 & 46.3 & 323 &  & 0.7 & 35.2 & 69.6 & 211 \\
     & 0.6 & 71.0 & 38.0 & 395 &  & 0.6 & 34.2 & 59.3 & 250 \\
     & 0.5 & 66.1 & 32.4 & 466 &  & 0.5 & 31.0 & 49.8 & 297 \\
    \midrule
    Learn2PD & 0.99 & 78.2 & 212.5 & 71 & Learn2PD & 0.99 & 36.2 & 223.3 & 66 \\
     & 0.98 & 77.6 & 90.0 & 163 &  & 0.98 & 37.2 & 117.7 & 124 \\
     & 0.96 & 76.9 & 61.2 & 242 &  & 0.96 & 38.0 & 87.5 & 164 \\
     & 0.94 & 76.0 & 51.8 & 282 &  & 0.94 & 37.4 & 76.9 & 188 \\
     & 0.9 & 73.6 & 43.1 & 342 &  & 0.9 & 36.4 & 65.8 & 218 \\
     & 0.85 & 73.9 & 37.8 & 397 &  & 0.85 & 33.4 & 58.7 & 249 \\
     & 0.8 & 70.4 & 34.8 & 426 &  & 0.8 & 30.6 & 53.9 & 271 \\
     & 0.7 & 66.2 & 31.9 & 467 &  & 0.7 & 29.0 & 48.8 & 297 \\
     & 0.6 & 59.4 & 29.5 & 506 &  & 0.6 & 24.6 & 43.9 & 331 \\
     & 0.5 & 53.1 & 27.5 & 539 &  & 0.5 & 24.0 & 40.1 & 357 \\
    \midrule
    EB-Sampler & 10 & 60.5 & 26.6 & 553 & EB-Sampler & 10 & 25.6 & 32.8 & 430 \\
     & 3 & 71.9 & 35.8 & 414 &  & 3 & 33.6 & 47.2 & 305 \\
     & 1 & 75.1 & 47.4 & 313 &  & 1 & 38.6 & 66.0 & 221 \\
     & 0.3 & 77.9 & 66.8 & 221 &  & 0.3 & 40.0 & 91.7 & 158 \\
     & 0.1 & 77.5 & 84.8 & 176 &  & 0.1 & 38.8 & 114.4 & 127 \\
     & 0.03 & 77.7 & 106.1 & 139 &  & 0.03 & 39.0 & 139.7 & 103 \\
     & 0.01 & 77.8 & 127.7 & 117 &  & 0.01 & 38.4 & 163.6 & 88 \\
     & 0.003 & 78.5 & 155.5 & 96 &  & 0.003 & 39.2 & 191.7 & 76 \\
     & 0.001 & 78.3 & 180.3 & 83 &  & 0.001 & 38.8 & 214.0 & 68 \\
     & 0 & 78.0 & 255.9 & 58 &  & 0 & 38.8 & 255.9 & 57 \\
    \bottomrule
\end{tabular}

%% file: tab/llada8b_humaneval_mbpp.tex
\begin{tabular}{ll rrr @{\hspace{1.5em}} ll rrr}
    \toprule
    \multicolumn{5}{c}{HumanEval} & \multicolumn{5}{c}{MBPP} \\
    \cmidrule(lr){1-5}\cmidrule(lr){6-10}
    Method & Param. & Acc. & NFE & TPS & Method & Param. & Acc. & NFE & TPS \\
    \midrule
    OnTrack (Ours) & 0.99 & 43.3 & 256.0 & 39 & OnTrack (Ours) & 0.99 & 53.7 & 256.0 & 38 \\
     & 0.98 & 43.3 & 247.1 & 40 &  & 0.98 & 53.7 & 196.9 & 48 \\
     & 0.96 & 43.3 & 212.9 & 46 &  & 0.96 & 53.7 & 131.9 & 72 \\
     & 0.94 & 43.9 & 176.5 & 57 &  & 0.94 & 53.7 & 103.3 & 93 \\
     & 0.9 & 43.3 & 130.0 & 76 &  & 0.9 & 54.1 & 73.4 & 129 \\
     & 0.85 & 42.1 & 99.1 & 101 &  & 0.85 & 54.1 & 57.1 & 163 \\
     & 0.8 & 41.5 & 82.9 & 120 &  & 0.8 & 52.5 & 48.6 & 193 \\
     & 0.7 & 35.4 & 65.6 & 153 &  & 0.7 & 45.9 & 42.4 & 237 \\
     & 0.6 & 33.5 & 54.8 & 180 &  & 0.6 & 42.0 & 39.4 & 264 \\
     & 0.5 & 25.6 & 44.0 & 215 &  & 0.5 & 37.0 & 35.9 & 293 \\
    \midrule
    Fast-dLLM & 0.99 & 45.1 & 163.9 & 63 & Fast-dLLM & 0.99 & 52.9 & 97.1 & 104 \\
     & 0.98 & 45.1 & 153.0 & 67 &  & 0.98 & 52.9 & 89.7 & 112 \\
     & 0.96 & 45.1 & 137.8 & 75 &  & 0.96 & 52.9 & 80.4 & 124 \\
     & 0.94 & 45.1 & 129.1 & 80 &  & 0.94 & 52.9 & 75.2 & 135 \\
     & 0.9 & 44.5 & 115.7 & 90 &  & 0.9 & 52.5 & 66.9 & 151 \\
     & 0.85 & 43.3 & 104.9 & 98 &  & 0.85 & 52.5 & 60.8 & 164 \\
     & 0.8 & 42.1 & 95.6 & 108 &  & 0.8 & 51.8 & 55.5 & 180 \\
     & 0.7 & 39.0 & 78.2 & 133 &  & 0.7 & 49.8 & 46.2 & 218 \\
     & 0.6 & 34.8 & 64.0 & 166 &  & 0.6 & 44.4 & 39.2 & 259 \\
     & 0.5 & 25.6 & 52.7 & 196 &  & 0.5 & 40.1 & 38.0 & 290 \\
    \midrule
    Learn2PD & 0.99 & 43.3 & 227.1 & 46 & Learn2PD & 0.99 & 53.3 & 199.0 & 50 \\
     & 0.98 & 42.7 & 132.5 & 80 &  & 0.98 & 53.3 & 76.0 & 129 \\
     & 0.96 & 41.5 & 101.7 & 104 &  & 0.96 & 52.9 & 57.2 & 172 \\
     & 0.94 & 38.4 & 88.9 & 118 &  & 0.94 & 52.5 & 50.8 & 196 \\
     & 0.9 & 36.0 & 75.6 & 139 &  & 0.9 & 48.6 & 44.9 & 225 \\
     & 0.85 & 33.5 & 65.0 & 156 &  & 0.85 & 48.2 & 42.1 & 246 \\
     & 0.8 & 31.7 & 57.2 & 179 &  & 0.8 & 43.6 & 41.7 & 263 \\
     & 0.7 & 20.7 & 49.8 & 199 &  & 0.7 & 32.7 & 39.7 & 283 \\
     & 0.6 & 17.1 & 43.2 & 230 &  & 0.6 & 28.4 & 36.7 & 300 \\
     & 0.5 & 12.8 & 36.1 & 266 &  & 0.5 & 24.1 & 32.3 & 334 \\
    \midrule
    EB-Sampler & 10 & 16.5 & 29.9 & 318 & EB-Sampler & 10 & 21.0 & 23.2 & 402 \\
     & 3 & 29.3 & 48.8 & 210 &  & 3 & 38.1 & 37.6 & 280 \\
     & 1 & 38.4 & 73.2 & 141 &  & 1 & 51.0 & 49.3 & 210 \\
     & 0.3 & 42.7 & 104.8 & 99 &  & 0.3 & 54.9 & 65.1 & 157 \\
     & 0.1 & 43.9 & 130.6 & 79 &  & 0.1 & 54.1 & 78.2 & 128 \\
     & 0.03 & 47.0 & 159.0 & 65 &  & 0.03 & 53.3 & 96.9 & 103 \\
     & 0.01 & 46.3 & 182.4 & 56 &  & 0.01 & 53.3 & 113.6 & 89 \\
     & 0.003 & 45.1 & 207.3 & 49 &  & 0.003 & 53.3 & 133.1 & 75 \\
     & 0.001 & 45.1 & 224.0 & 46 &  & 0.001 & 53.3 & 150.4 & 67 \\
     & 0 & 45.1 & 256.0 & 39 &  & 0 & 52.9 & 256.0 & 39 \\
    \bottomrule
\end{tabular}

%% file: tab/prefix_cache_gsm8k_zero_shot_full.tex
\begin{tabular}{ll rrr @{\hspace{1.5em}} ll rrr}
    \toprule
    \multicolumn{5}{c}{No cache} & \multicolumn{5}{c}{Prefix cache} \\
    \cmidrule(lr){1-5}\cmidrule(lr){6-10}
    Method & Param. & Acc. & NFE & TPS & Method & Param. & Acc. & NFE & TPS \\
    \midrule
    OnTrack (Ours) & 0.99 & 78.5 & 256.0 & 54 & OnTrack (Ours) & 0.99 & 77.3 & 256.0 & 58 \\
     & 0.98 & 78.4 & 109.0 & 127 &  & 0.98 & 77.3 & 111.5 & 132 \\
     & 0.96 & 78.8 & 70.6 & 196 &  & 0.96 & 78.2 & 72.0 & 203 \\
     & 0.94 & 77.9 & 60.6 & 230 &  & 0.94 & 77.9 & 61.9 & 239 \\
     & 0.9 & 77.8 & 51.7 & 274 &  & 0.9 & 77.9 & 52.7 & 249 \\
     & 0.85 & 76.3 & 45.8 & 308 &  & 0.85 & 75.3 & 46.5 & 311 \\
     & 0.8 & 77.0 & 42.1 & 331 &  & 0.8 & 74.7 & 42.7 & 324 \\
     & 0.7 & 75.7 & 37.1 & 381 &  & 0.7 & 73.5 & 37.8 & 387 \\
     & 0.6 & 73.9 & 34.2 & 411 &  & 0.6 & 74.5 & 34.7 & 423 \\
     & 0.5 & 71.9 & 31.8 & 440 &  & 0.5 & 73.8 & 32.1 & 453 \\
    \midrule
    Fast-dLLM & 0.99 & 78.2 & 118.1 & 128 & Fast-dLLM & 0.99 & 78.8 & 120.4 & 136 \\
     & 0.98 & 78.3 & 107.9 & 140 &  & 0.98 & 78.8 & 110.1 & 146 \\
     & 0.96 & 78.2 & 94.5 & 160 &  & 0.96 & 78.8 & 96.4 & 168 \\
     & 0.94 & 77.9 & 86.3 & 174 &  & 0.94 & 78.8 & 88.1 & 186 \\
     & 0.9 & 77.5 & 74.4 & 202 &  & 0.9 & 78.8 & 76.1 & 216 \\
     & 0.85 & 77.0 & 65.6 & 232 &  & 0.85 & 77.7 & 67.0 & 245 \\
     & 0.8 & 76.1 & 58.2 & 258 &  & 0.8 & 76.8 & 59.3 & 276 \\
     & 0.7 & 74.7 & 46.3 & 323 &  & 0.7 & 75.4 & 47.2 & 349 \\
     & 0.6 & 71.0 & 38.0 & 395 &  & 0.6 & 73.1 & 39.0 & 420 \\
     & 0.5 & 66.1 & 32.4 & 466 &  & 0.5 & 65.2 & 33.1 & 487 \\
    \midrule
    Learn2PD & 0.99 & 78.2 & 212.5 & 71 & Learn2PD & 0.99 & 77.6 & 214.1 & 75 \\
     & 0.98 & 77.6 & 90.0 & 163 &  & 0.98 & 78.2 & 90.9 & 177 \\
     & 0.96 & 76.9 & 61.2 & 242 &  & 0.96 & 77.7 & 62.6 & 253 \\
     & 0.94 & 76.0 & 51.8 & 282 &  & 0.94 & 75.5 & 53.1 & 299 \\
     & 0.9 & 73.6 & 43.1 & 342 &  & 0.9 & 74.1 & 44.3 & 359 \\
     & 0.85 & 73.9 & 37.8 & 397 &  & 0.85 & 72.3 & 39.0 & 406 \\
     & 0.8 & 70.4 & 34.8 & 426 &  & 0.8 & 70.2 & 35.9 & 442 \\
     & 0.7 & 66.2 & 31.9 & 467 &  & 0.7 & 64.8 & 32.3 & 490 \\
     & 0.6 & 59.4 & 29.5 & 506 &  & 0.6 & 63.2 & 29.9 & 523 \\
     & 0.5 & 53.1 & 27.5 & 539 &  & 0.5 & 54.7 & 28.2 & 556 \\
    \midrule
    EB-Sampler & 10 & 60.5 & 26.6 & 553 & EB-Sampler & 10 & 60.0 & 27.0 & 568 \\
     & 3 & 71.9 & 35.8 & 414 &  & 3 & 73.1 & 36.1 & 432 \\
     & 1 & 75.1 & 47.4 & 313 &  & 1 & 76.3 & 48.1 & 329 \\
     & 0.3 & 77.9 & 66.8 & 221 &  & 0.3 & 76.7 & 68.1 & 234 \\
     & 0.1 & 77.5 & 84.8 & 176 &  & 0.1 & 77.2 & 87.1 & 183 \\
     & 0.03 & 77.7 & 106.1 & 139 &  & 0.03 & 78.1 & 108.2 & 148 \\
     & 0.01 & 77.8 & 127.7 & 117 &  & 0.01 & 78.0 & 129.7 & 123 \\
     & 0.003 & 78.5 & 155.5 & 96 &  & 0.003 & 78.5 & 157.0 & 104 \\
     & 0.001 & 78.3 & 180.3 & 83 &  & 0.001 & 78.8 & 181.3 & 89 \\
     & 0 & 78.0 & 255.9 & 58 &  & 0 & 78.8 & 255.9 & 62 \\
    \bottomrule
\end{tabular}

%% file: tab/prefix_cache_gsm8k_four_shot_full.tex
\begin{tabular}{ll rrr @{\hspace{1.5em}} ll rrr}
    \toprule
    \multicolumn{5}{c}{No cache} & \multicolumn{5}{c}{Prefix cache} \\
    \cmidrule(lr){1-5}\cmidrule(lr){6-10}
    Method & Param. & Acc. & NFE & TPS & Method & Param. & Acc. & NFE & TPS \\
    \midrule
    OnTrack (Ours) & 0.99 & 80.8 & 256.0 & 22 & OnTrack (Ours) & 0.99 & 79.3 & 256.0 & 50 \\
     & 0.98 & 81.1 & 112.6 & 50 &  & 0.98 & 79.4 & 117.3 & 106 \\
     & 0.96 & 80.9 & 72.6 & 78 &  & 0.96 & 78.8 & 76.1 & 154 \\
     & 0.94 & 80.5 & 62.1 & 91 &  & 0.94 & 78.8 & 65.1 & 178 \\
     & 0.9 & 79.2 & 52.8 & 107 &  & 0.9 & 77.2 & 55.1 & 203 \\
     & 0.85 & 78.7 & 46.9 & 121 &  & 0.85 & 78.5 & 48.7 & 225 \\
     & 0.8 & 77.5 & 42.8 & 132 &  & 0.8 & 76.9 & 44.8 & 241 \\
     & 0.7 & 76.9 & 37.9 & 149 &  & 0.7 & 75.9 & 39.5 & 267 \\
     & 0.6 & 75.8 & 34.9 & 162 &  & 0.6 & 74.4 & 36.3 & 288 \\
     & 0.5 & 71.9 & 32.9 & 171 &  & 0.5 & 72.4 & 33.5 & 302 \\
    \midrule
    Fast-dLLM & 0.99 & 80.1 & 122.3 & 47 & Fast-dLLM & 0.99 & 80.4 & 126.9 & 103 \\
     & 0.98 & 80.1 & 111.8 & 52 &  & 0.98 & 80.4 & 116.2 & 111 \\
     & 0.96 & 80.1 & 97.9 & 59 &  & 0.96 & 80.7 & 101.9 & 125 \\
     & 0.94 & 80.1 & 89.1 & 65 &  & 0.94 & 80.4 & 93.2 & 135 \\
     & 0.9 & 80.0 & 76.5 & 76 &  & 0.9 & 79.8 & 80.5 & 155 \\
     & 0.85 & 80.3 & 67.2 & 86 &  & 0.85 & 79.8 & 70.6 & 173 \\
     & 0.8 & 79.8 & 59.2 & 98 &  & 0.8 & 78.8 & 62.2 & 193 \\
     & 0.7 & 78.0 & 46.8 & 122 &  & 0.7 & 76.5 & 49.4 & 234 \\
     & 0.6 & 72.5 & 38.7 & 149 &  & 0.6 & 74.1 & 40.6 & 274 \\
     & 0.5 & 65.7 & 33.3 & 172 &  & 0.5 & 65.7 & 34.4 & 316 \\
    \midrule
    Learn2PD & 0.99 & 79.8 & 213.1 & 27 & Learn2PD & 0.99 & 80.2 & 214.9 & 63 \\
     & 0.98 & 79.5 & 89.8 & 64 &  & 0.98 & 79.9 & 93.9 & 134 \\
     & 0.96 & 79.5 & 62.1 & 92 &  & 0.96 & 79.3 & 65.4 & 183 \\
     & 0.94 & 77.9 & 52.4 & 110 &  & 0.94 & 78.2 & 55.0 & 212 \\
     & 0.9 & 75.9 & 43.7 & 131 &  & 0.9 & 76.3 & 46.0 & 247 \\
     & 0.85 & 72.9 & 38.6 & 149 &  & 0.85 & 72.4 & 40.4 & 275 \\
     & 0.8 & 70.1 & 35.8 & 161 &  & 0.8 & 72.3 & 37.2 & 293 \\
     & 0.7 & 62.7 & 32.9 & 174 &  & 0.7 & 66.4 & 33.8 & 313 \\
     & 0.6 & 58.0 & 30.8 & 187 &  & 0.6 & 59.8 & 31.6 & 329 \\
     & 0.5 & 53.4 & 29.0 & 198 &  & 0.5 & 53.3 & 29.8 & 346 \\
    \midrule
    EB-Sampler & 10 & 61.6 & 27.9 & 205 & EB-Sampler & 10 & 61.9 & 28.5 & 355 \\
     & 3 & 74.6 & 36.6 & 156 &  & 3 & 74.5 & 37.6 & 288 \\
     & 1 & 78.4 & 47.9 & 120 &  & 1 & 78.2 & 50.2 & 231 \\
     & 0.3 & 80.0 & 69.0 & 82 &  & 0.3 & 79.4 & 71.7 & 171 \\
     & 0.1 & 80.4 & 87.1 & 65 &  & 0.1 & 80.2 & 90.9 & 137 \\
     & 0.03 & 80.1 & 108.6 & 52 &  & 0.03 & 80.1 & 112.7 & 114 \\
     & 0.01 & 79.8 & 129.2 & 44 &  & 0.01 & 80.0 & 133.3 & 98 \\
     & 0.003 & 80.0 & 155.6 & 36 &  & 0.003 & 79.8 & 159.5 & 83 \\
     & 0.001 & 79.6 & 179.0 & 32 &  & 0.001 & 80.4 & 182.3 & 73 \\
     & 0 & 79.9 & 255.7 & 23 &  & 0 & 80.2 & 255.9 & 54 \\
    \bottomrule
\end{tabular}

%% file: tab/dream7b_gsm8k_math.tex
\begin{tabular}{ll rrr @{\hspace{1.5em}} ll rrr}
    \toprule
    \multicolumn{5}{c}{GSM8K} & \multicolumn{5}{c}{MATH-500} \\
    \cmidrule(lr){1-5}\cmidrule(lr){6-10}
    Method & Param. & Acc. & NFE & TPS & Method & Param. & Acc. & NFE & TPS \\
    \midrule
    OnTrack (Ours) & 0.99 & 81.7 & 256.0 & 52 & OnTrack (Ours) & 0.99 & 46.0 & 256.0 & 54 \\
     & 0.98 & 81.6 & 154.4 & 85 &  & 0.98 & 46.0 & 158.7 & 86 \\
     & 0.96 & 81.5 & 98.4 & 131 &  & 0.96 & 46.6 & 103.5 & 132 \\
     & 0.94 & 81.2 & 83.7 & 154 &  & 0.94 & 46.8 & 89.7 & 152 \\
     & 0.9 & 80.4 & 70.6 & 184 &  & 0.9 & 46.0 & 77.0 & 175 \\
     & 0.85 & 78.4 & 62.4 & 206 &  & 0.85 & 45.8 & 69.1 & 194 \\
     & 0.8 & 78.6 & 57.5 & 223 &  & 0.8 & 43.0 & 64.0 & 210 \\
     & 0.7 & 75.5 & 50.6 & 255 &  & 0.7 & 40.6 & 57.0 & 241 \\
     & 0.6 & 71.6 & 46.5 & 284 &  & 0.6 & 36.2 & 51.8 & 265 \\
     & 0.5 & 65.7 & 42.9 & 305 &  & 0.5 & 32.2 & 46.7 & 290 \\
    \midrule
    Fast-dLLM & 0.99 & 83.0 & 140.2 & 102 & Fast-dLLM & 0.99 & 47.2 & 155.1 & 95 \\
     & 0.98 & 83.0 & 130.3 & 108 &  & 0.98 & 47.2 & 144.1 & 100 \\
     & 0.96 & 82.9 & 116.8 & 122 &  & 0.96 & 47.0 & 129.7 & 112 \\
     & 0.94 & 82.5 & 108.4 & 130 &  & 0.94 & 47.0 & 121.0 & 119 \\
     & 0.9 & 82.1 & 95.8 & 146 &  & 0.9 & 47.4 & 108.2 & 135 \\
     & 0.85 & 81.8 & 86.2 & 163 &  & 0.85 & 46.0 & 98.1 & 148 \\
     & 0.8 & 79.2 & 77.4 & 180 &  & 0.8 & 45.6 & 89.8 & 163 \\
     & 0.7 & 75.4 & 63.8 & 215 &  & 0.7 & 43.4 & 75.7 & 193 \\
     & 0.6 & 66.3 & 53.7 & 264 &  & 0.6 & 38.2 & 64.2 & 230 \\
     & 0.5 & 53.6 & 45.4 & 312 &  & 0.5 & 32.8 & 54.2 & 271 \\
    \midrule
    Learn2PD & 0.99 & 82.1 & 212.5 & 65 & Learn2PD & 0.99 & 44.0 & 219.7 & 66 \\
     & 0.98 & 81.7 & 109.0 & 126 &  & 0.98 & 44.2 & 124.2 & 118 \\
     & 0.96 & 80.4 & 79.0 & 170 &  & 0.96 & 43.0 & 92.9 & 156 \\
     & 0.94 & 78.2 & 68.5 & 202 &  & 0.94 & 43.2 & 81.6 & 176 \\
     & 0.9 & 74.6 & 58.6 & 230 &  & 0.9 & 42.4 & 71.3 & 201 \\
     & 0.85 & 69.1 & 52.5 & 263 &  & 0.85 & 38.8 & 64.1 & 226 \\
     & 0.8 & 61.6 & 48.4 & 286 &  & 0.8 & 35.2 & 58.7 & 246 \\
     & 0.7 & 54.5 & 43.1 & 316 &  & 0.7 & 32.2 & 51.9 & 275 \\
     & 0.6 & 44.0 & 39.7 & 351 &  & 0.6 & 26.4 & 46.9 & 305 \\
     & 0.5 & 36.3 & 36.4 & 382 &  & 0.5 & 20.8 & 43.0 & 334 \\
    \midrule
    EB-Sampler & 10 & 34.7 & 32.2 & 438 & EB-Sampler & 10 & 19.6 & 34.0 & 428 \\
     & 3 & 65.7 & 44.9 & 307 &  & 3 & 32.4 & 50.0 & 288 \\
     & 1 & 76.1 & 59.1 & 235 &  & 1 & 41.0 & 68.7 & 211 \\
     & 0.3 & 80.3 & 83.6 & 165 &  & 0.3 & 46.0 & 96.2 & 147 \\
     & 0.1 & 81.4 & 103.6 & 134 &  & 0.1 & 45.6 & 118.6 & 119 \\
     & 0.03 & 82.3 & 123.7 & 111 &  & 0.03 & 48.0 & 142.1 & 101 \\
     & 0.01 & 82.8 & 141.7 & 96 &  & 0.01 & 47.0 & 162.8 & 88 \\
     & 0.003 & 83.2 & 162.8 & 84 &  & 0.003 & 47.0 & 185.8 & 78 \\
     & 0.001 & 83.0 & 181.4 & 76 &  & 0.001 & 46.8 & 204.4 & 71 \\
     & 0 & 83.1 & 256.0 & 54 &  & 0 & 47.2 & 256.0 & 57 \\
    \bottomrule
\end{tabular}

%% file: tab/dream7b_humaneval_mbpp.tex
\begin{tabular}{ll rrr @{\hspace{1.5em}} ll rrr}
    \toprule
    \multicolumn{5}{c}{HumanEval} & \multicolumn{5}{c}{MBPP} \\
    \cmidrule(lr){1-5}\cmidrule(lr){6-10}
    Method & Param. & Acc. & NFE & TPS & Method & Param. & Acc. & NFE & TPS \\
    \midrule
    OnTrack (Ours) & 0.99 & 58.5 & 256.0 & 27 & OnTrack (Ours) & 0.99 & 66.9 & 256.0 & 12 \\
     & 0.98 & 58.5 & 140.5 & 49 &  & 0.98 & 66.9 & 65.7 & 44 \\
     & 0.96 & 58.5 & 121.3 & 56 &  & 0.96 & 66.9 & 54.6 & 52 \\
     & 0.94 & 58.5 & 104.6 & 65 &  & 0.94 & 66.9 & 47.3 & 61 \\
     & 0.9 & 59.8 & 84.1 & 81 &  & 0.9 & 65.8 & 38.5 & 73 \\
     & 0.85 & 56.7 & 69.8 & 98 &  & 0.85 & 65.0 & 32.7 & 87 \\
     & 0.8 & 55.5 & 61.2 & 111 &  & 0.8 & 62.3 & 30.2 & 98 \\
     & 0.7 & 48.8 & 50.0 & 133 &  & 0.7 & 61.1 & 26.3 & 120 \\
     & 0.6 & 40.9 & 43.6 & 152 &  & 0.6 & 55.6 & 22.8 & 141 \\
     & 0.5 & 35.4 & 38.2 & 174 &  & 0.5 & 49.0 & 21.2 & 159 \\
    \midrule
    Fast-dLLM & 0.99 & 55.5 & 109.9 & 64 & Fast-dLLM & 0.99 & 65.8 & 45.9 & 65 \\
     & 0.98 & 55.5 & 104.5 & 67 &  & 0.98 & 65.8 & 43.3 & 69 \\
     & 0.96 & 55.5 & 97.1 & 70 &  & 0.96 & 65.8 & 38.9 & 76 \\
     & 0.94 & 54.9 & 91.8 & 77 &  & 0.94 & 65.8 & 36.1 & 82 \\
     & 0.9 & 53.7 & 83.7 & 86 &  & 0.9 & 65.4 & 32.4 & 94 \\
     & 0.85 & 52.4 & 76.9 & 93 &  & 0.85 & 65.0 & 29.4 & 103 \\
     & 0.8 & 50.6 & 70.2 & 100 &  & 0.8 & 63.8 & 27.1 & 112 \\
     & 0.7 & 50.0 & 59.0 & 120 &  & 0.7 & 60.7 & 23.8 & 128 \\
     & 0.6 & 43.3 & 48.8 & 141 &  & 0.6 & 52.9 & 22.1 & 149 \\
     & 0.5 & 29.3 & 41.0 & 169 &  & 0.5 & 44.4 & 20.4 & 169 \\
    \midrule
    Learn2PD & 0.99 & 57.3 & 199.7 & 34 & Learn2PD & 0.99 & 66.5 & 180.0 & 16 \\
     & 0.98 & 56.1 & 92.3 & 75 &  & 0.98 & 65.8 & 41.6 & 70 \\
     & 0.96 & 54.3 & 68.8 & 99 &  & 0.96 & 65.4 & 28.5 & 104 \\
     & 0.94 & 52.4 & 61.2 & 110 &  & 0.94 & 64.6 & 25.8 & 116 \\
     & 0.9 & 47.0 & 52.5 & 127 &  & 0.9 & 61.1 & 24.7 & 126 \\
     & 0.85 & 40.2 & 47.2 & 142 &  & 0.85 & 57.6 & 22.8 & 143 \\
     & 0.8 & 33.5 & 42.2 & 157 &  & 0.8 & 48.6 & 23.3 & 150 \\
     & 0.7 & 22.6 & 36.7 & 182 &  & 0.7 & 42.8 & 21.9 & 178 \\
     & 0.6 & 14.0 & 33.0 & 207 &  & 0.6 & 38.9 & 19.5 & 207 \\
     & 0.5 & 11.0 & 29.4 & 227 &  & 0.5 & 27.2 & 17.4 & 204 \\
    \midrule
    EB-Sampler & 10 & 12.2 & 23.2 & 289 & EB-Sampler & 10 & 25.7 & 15.0 & 233 \\
     & 3 & 36.6 & 36.3 & 187 &  & 3 & 45.5 & 23.0 & 169 \\
     & 1 & 48.2 & 52.1 & 131 &  & 1 & 59.5 & 30.2 & 125 \\
     & 0.3 & 53.0 & 71.4 & 91 &  & 0.3 & 66.9 & 32.5 & 96 \\
     & 0.1 & 53.7 & 85.7 & 76 &  & 0.1 & 66.5 & 36.9 & 77 \\
     & 0.03 & 53.7 & 100.2 & 67 &  & 0.03 & 66.1 & 43.5 & 67 \\
     & 0.01 & 54.3 & 112.0 & 63 &  & 0.01 & 65.4 & 50.2 & 60 \\
     & 0.003 & 55.5 & 124.5 & 57 &  & 0.003 & 65.8 & 58.6 & 51 \\
     & 0.001 & 55.5 & 134.1 & 53 &  & 0.001 & 65.8 & 66.0 & 45 \\
     & 0 & 55.5 & 256.0 & 27 &  & 0 & 65.8 & 256.0 & 12 \\
    \bottomrule
\end{tabular}

%% file: tab/llada8b_length128_gsm8k_math.tex
\begin{tabular}{ll rrr @{\hspace{1.5em}} ll rrr}
    \toprule
    \multicolumn{5}{c}{GSM8K} & \multicolumn{5}{c}{MATH-500} \\
    \cmidrule(lr){1-5}\cmidrule(lr){6-10}
    Method & Param. & Acc. & NFE & TPS & Method & Param. & Acc. & NFE & TPS \\
    \midrule
    OnTrack (Ours) & 0.99 & 72.5 & 128.0 & 64 & OnTrack (Ours) & 0.99 & 35.0 & 128.0 & 58 \\
     & 0.98 & 72.3 & 67.3 & 124 &  & 0.98 & 35.0 & 86.2 & 92 \\
     & 0.96 & 72.3 & 45.8 & 181 &  & 0.96 & 35.2 & 57.8 & 128 \\
     & 0.94 & 71.9 & 39.7 & 210 &  & 0.94 & 33.0 & 50.2 & 153 \\
     & 0.9 & 72.5 & 34.1 & 245 &  & 0.9 & 34.0 & 43.3 & 181 \\
     & 0.85 & 71.7 & 30.3 & 273 &  & 0.85 & 34.0 & 38.6 & 199 \\
     & 0.8 & 70.2 & 27.9 & 298 &  & 0.8 & 33.6 & 35.5 & 215 \\
     & 0.7 & 69.7 & 24.6 & 327 &  & 0.7 & 34.4 & 31.1 & 230 \\
     & 0.6 & 67.8 & 22.4 & 372 &  & 0.6 & 31.0 & 28.1 & 266 \\
     & 0.5 & 67.1 & 20.7 & 403 &  & 0.5 & 27.0 & 26.1 & 299 \\
    \midrule
    Fast-dLLM & 0.99 & 73.0 & 76.8 & 118 & Fast-dLLM & 0.99 & 34.8 & 91.5 & 94 \\
     & 0.98 & 73.1 & 70.5 & 128 &  & 0.98 & 34.8 & 85.4 & 99 \\
     & 0.96 & 73.4 & 62.4 & 147 &  & 0.96 & 34.8 & 77.0 & 108 \\
     & 0.94 & 73.3 & 57.5 & 159 &  & 0.94 & 35.2 & 71.8 & 117 \\
     & 0.9 & 73.4 & 50.4 & 181 &  & 0.9 & 35.6 & 64.2 & 133 \\
     & 0.85 & 73.2 & 44.9 & 203 &  & 0.85 & 35.0 & 58.3 & 146 \\
     & 0.8 & 73.5 & 40.0 & 228 &  & 0.8 & 34.0 & 53.1 & 161 \\
     & 0.7 & 71.0 & 32.4 & 283 &  & 0.7 & 31.8 & 44.1 & 193 \\
     & 0.6 & 69.8 & 26.6 & 347 &  & 0.6 & 31.8 & 37.4 & 221 \\
     & 0.5 & 62.4 & 21.9 & 423 &  & 0.5 & 27.4 & 30.8 & 280 \\
    \midrule
    Learn2PD & 0.99 & 73.1 & 112.0 & 79 & Learn2PD & 0.99 & 33.2 & 115.5 & 75 \\
     & 0.98 & 73.2 & 59.9 & 149 &  & 0.98 & 33.6 & 72.4 & 118 \\
     & 0.96 & 72.6 & 42.6 & 209 &  & 0.96 & 33.0 & 54.8 & 157 \\
     & 0.94 & 72.8 & 36.4 & 246 &  & 0.94 & 32.2 & 48.1 & 179 \\
     & 0.9 & 70.7 & 30.5 & 298 &  & 0.9 & 31.0 & 41.6 & 204 \\
     & 0.85 & 67.2 & 26.6 & 343 &  & 0.85 & 30.0 & 36.7 & 233 \\
     & 0.8 & 66.0 & 24.3 & 375 &  & 0.8 & 27.8 & 33.5 & 254 \\
     & 0.7 & 61.0 & 21.4 & 419 &  & 0.7 & 28.6 & 29.1 & 293 \\
     & 0.6 & 55.3 & 19.4 & 461 &  & 0.6 & 24.6 & 26.4 & 317 \\
     & 0.5 & 49.3 & 17.9 & 498 &  & 0.5 & 20.6 & 23.8 & 357 \\
    \midrule
    EB-Sampler & 10 & 54.1 & 15.6 & 559 & EB-Sampler & 10 & 19.0 & 18.4 & 441 \\
     & 3 & 66.5 & 22.2 & 404 &  & 3 & 29.4 & 27.4 & 303 \\
     & 1 & 70.0 & 31.2 & 290 &  & 1 & 33.6 & 40.5 & 209 \\
     & 0.3 & 73.3 & 45.3 & 188 &  & 0.3 & 35.2 & 58.4 & 140 \\
     & 0.1 & 73.2 & 57.3 & 150 &  & 0.1 & 34.6 & 71.9 & 114 \\
     & 0.03 & 72.8 & 70.5 & 122 &  & 0.03 & 34.2 & 85.9 & 89 \\
     & 0.01 & 73.1 & 82.6 & 105 &  & 0.01 & 35.2 & 97.1 & 87 \\
     & 0.003 & 72.9 & 96.6 & 89 &  & 0.003 & 34.2 & 108.2 & 78 \\
     & 0.001 & 73.1 & 107.1 & 82 &  & 0.001 & 34.4 & 115.2 & 74 \\
     & 0 & 73.0 & 128.0 & 70 &  & 0 & 34.8 & 128.0 & 66 \\
    \bottomrule
\end{tabular}

%% file: tab/llada8b_length128_humaneval_mbpp.tex
\begin{tabular}{ll rrr @{\hspace{1.5em}} ll rrr}
    \toprule
    \multicolumn{5}{c}{HumanEval} & \multicolumn{5}{c}{MBPP} \\
    \cmidrule(lr){1-5}\cmidrule(lr){6-10}
    Method & Param. & Acc. & NFE & TPS & Method & Param. & Acc. & NFE & TPS \\
    \midrule
    OnTrack (Ours) & 0.99 & 43.9 & 128.0 & 53 & OnTrack (Ours) & 0.99 & 51.4 & 128.0 & 47 \\
     & 0.98 & 43.9 & 126.3 & 53 &  & 0.98 & 51.4 & 110.3 & 58 \\
     & 0.96 & 43.9 & 117.2 & 59 &  & 0.96 & 51.4 & 87.4 & 73 \\
     & 0.94 & 43.9 & 101.7 & 67 &  & 0.94 & 51.4 & 71.6 & 86 \\
     & 0.9 & 43.3 & 78.7 & 87 &  & 0.9 & 50.6 & 51.6 & 113 \\
     & 0.85 & 42.1 & 62.1 & 111 &  & 0.85 & 49.4 & 41.2 & 142 \\
     & 0.8 & 41.5 & 52.5 & 130 &  & 0.8 & 48.6 & 33.9 & 185 \\
     & 0.7 & 39.6 & 40.6 & 168 &  & 0.7 & 45.5 & 28.1 & 228 \\
     & 0.6 & 32.9 & 35.0 & 193 &  & 0.6 & 42.4 & 24.5 & 253 \\
     & 0.5 & 25.0 & 30.4 & 223 &  & 0.5 & 35.4 & 20.4 & 291 \\
    \midrule
    Fast-dLLM & 0.99 & 45.1 & 97.3 & 75 & Fast-dLLM & 0.99 & 51.8 & 70.0 & 98 \\
     & 0.98 & 45.1 & 91.0 & 80 &  & 0.98 & 51.8 & 65.4 & 104 \\
     & 0.96 & 44.5 & 83.0 & 87 &  & 0.96 & 51.8 & 59.0 & 113 \\
     & 0.94 & 44.5 & 78.0 & 92 &  & 0.94 & 51.8 & 55.1 & 121 \\
     & 0.9 & 44.5 & 70.2 & 104 &  & 0.9 & 52.5 & 48.7 & 137 \\
     & 0.85 & 43.3 & 63.5 & 115 &  & 0.85 & 51.0 & 44.0 & 154 \\
     & 0.8 & 43.9 & 57.8 & 126 &  & 0.8 & 51.0 & 39.3 & 171 \\
     & 0.7 & 42.7 & 47.9 & 151 &  & 0.7 & 47.9 & 33.4 & 204 \\
     & 0.6 & 38.4 & 39.4 & 182 &  & 0.6 & 44.7 & 28.5 & 240 \\
     & 0.5 & 28.0 & 33.1 & 219 &  & 0.5 & 37.4 & 24.1 & 287 \\
    \midrule
    Learn2PD & 0.99 & 41.5 & 117.2 & 62 & Learn2PD & 0.99 & 52.9 & 106.2 & 64 \\
     & 0.98 & 40.9 & 78.6 & 92 &  & 0.98 & 53.3 & 54.8 & 122 \\
     & 0.96 & 40.9 & 62.7 & 116 &  & 0.96 & 54.1 & 40.7 & 165 \\
     & 0.94 & 42.1 & 55.9 & 129 &  & 0.94 & 52.5 & 36.3 & 187 \\
     & 0.9 & 39.6 & 47.5 & 154 &  & 0.9 & 50.6 & 31.1 & 217 \\
     & 0.85 & 36.6 & 40.6 & 178 &  & 0.85 & 45.5 & 28.4 & 241 \\
     & 0.8 & 31.1 & 36.7 & 198 &  & 0.8 & 42.4 & 26.1 & 265 \\
     & 0.7 & 25.6 & 31.3 & 228 &  & 0.7 & 35.8 & 23.5 & 301 \\
     & 0.6 & 20.7 & 27.7 & 258 &  & 0.6 & 30.4 & 21.1 & 326 \\
     & 0.5 & 14.0 & 23.8 & 292 &  & 0.5 & 22.6 & 18.9 & 361 \\
    \midrule
    EB-Sampler & 10 & 15.2 & 18.0 & 347 & EB-Sampler & 10 & 26.5 & 14.5 & 429 \\
     & 3 & 28.7 & 30.2 & 236 &  & 3 & 38.5 & 22.5 & 294 \\
     & 1 & 36.0 & 42.5 & 168 &  & 1 & 47.9 & 32.3 & 212 \\
     & 0.3 & 43.9 & 62.3 & 115 &  & 0.3 & 51.4 & 45.0 & 144 \\
     & 0.1 & 43.9 & 76.6 & 93 &  & 0.1 & 51.4 & 55.7 & 111 \\
     & 0.03 & 44.5 & 90.8 & 78 &  & 0.03 & 51.8 & 67.0 & 99 \\
     & 0.01 & 44.5 & 101.6 & 71 &  & 0.01 & 51.8 & 76.2 & 89 \\
     & 0.003 & 45.1 & 111.5 & 64 &  & 0.003 & 52.1 & 85.8 & 79 \\
     & 0.001 & 45.1 & 117.4 & 61 &  & 0.001 & 51.8 & 92.8 & 73 \\
     & 0 & 45.1 & 128.0 & 56 &  & 0 & 51.8 & 128.0 & 54 \\
    \bottomrule
\end{tabular}

%% file: tab/llada8b_length512_gsm8k_math.tex
\begin{tabular}{ll rrr @{\hspace{1.5em}} ll rrr}
    \toprule
    \multicolumn{5}{c}{GSM8K} & \multicolumn{5}{c}{MATH-500} \\
    \cmidrule(lr){1-5}\cmidrule(lr){6-10}
    Method & Param. & Acc. & NFE & TPS & Method & Param. & Acc. & NFE & TPS \\
    \midrule
    OnTrack (Ours) & 0.99 & 82.2 & 348.5 & 39 & OnTrack (Ours) & 0.99 & 39.0 & 445.9 & 40 \\
     & 0.98 & 82.4 & 136.2 & 102 &  & 0.98 & 39.4 & 208.3 & 85 \\
     & 0.96 & 82.9 & 86.6 & 160 &  & 0.96 & 38.4 & 134.9 & 130 \\
     & 0.94 & 82.3 & 74.3 & 186 &  & 0.94 & 37.6 & 117.0 & 150 \\
     & 0.9 & 82.3 & 63.1 & 217 &  & 0.9 & 39.8 & 101.1 & 174 \\
     & 0.85 & 80.9 & 55.9 & 246 &  & 0.85 & 39.2 & 89.6 & 196 \\
     & 0.8 & 79.9 & 51.5 & 267 &  & 0.8 & 38.4 & 82.6 & 214 \\
     & 0.7 & 78.5 & 45.5 & 302 &  & 0.7 & 37.6 & 73.3 & 237 \\
     & 0.6 & 77.3 & 42.1 & 323 &  & 0.6 & 35.8 & 67.0 & 262 \\
     & 0.5 & 73.6 & 39.2 & 343 &  & 0.5 & 34.6 & 62.0 & 283 \\
    \midrule
    Fast-dLLM & 0.99 & 82.3 & 149.0 & 99 & Fast-dLLM & 0.99 & 41.4 & 228.1 & 82 \\
     & 0.98 & 82.0 & 135.5 & 107 &  & 0.98 & 41.8 & 208.6 & 90 \\
     & 0.96 & 82.0 & 117.9 & 123 &  & 0.96 & 40.8 & 184.8 & 101 \\
     & 0.94 & 82.2 & 107.3 & 135 &  & 0.94 & 43.0 & 169.6 & 110 \\
     & 0.9 & 81.7 & 92.4 & 158 &  & 0.9 & 41.4 & 148.5 & 125 \\
     & 0.85 & 81.6 & 81.2 & 181 &  & 0.85 & 41.6 & 133.0 & 139 \\
     & 0.8 & 81.9 & 71.6 & 202 &  & 0.8 & 39.2 & 120.3 & 153 \\
     & 0.7 & 81.0 & 57.1 & 255 &  & 0.7 & 39.2 & 98.1 & 189 \\
     & 0.6 & 77.7 & 47.4 & 304 &  & 0.6 & 34.0 & 83.2 & 222 \\
     & 0.5 & 69.8 & 40.5 & 356 &  & 0.5 & 29.8 & 72.2 & 257 \\
    \midrule
    Learn2PD & 0.99 & 81.8 & 284.0 & 50 & Learn2PD & 0.99 & 42.0 & 378.1 & 48 \\
     & 0.98 & 81.3 & 113.3 & 128 &  & 0.98 & 42.2 & 173.9 & 104 \\
     & 0.96 & 81.6 & 75.7 & 189 &  & 0.96 & 40.6 & 125.3 & 145 \\
     & 0.94 & 82.0 & 63.8 & 224 &  & 0.94 & 39.0 & 107.7 & 168 \\
     & 0.9 & 79.1 & 53.4 & 266 &  & 0.9 & 38.0 & 92.2 & 199 \\
     & 0.85 & 77.0 & 48.0 & 295 &  & 0.85 & 34.2 & 82.8 & 222 \\
     & 0.8 & 74.5 & 44.2 & 318 &  & 0.8 & 35.4 & 76.2 & 242 \\
     & 0.7 & 69.4 & 40.1 & 352 &  & 0.7 & 32.4 & 69.7 & 264 \\
     & 0.6 & 63.3 & 37.3 & 380 &  & 0.6 & 27.4 & 63.8 & 286 \\
     & 0.5 & 56.9 & 34.8 & 409 &  & 0.5 & 21.4 & 58.1 & 315 \\
    \midrule
    EB-Sampler & 10 & 60.7 & 35.1 & 403 & EB-Sampler & 10 & 20.4 & 52.1 & 344 \\
     & 3 & 75.4 & 45.6 & 307 &  & 3 & 34.8 & 73.1 & 245 \\
     & 1 & 79.5 & 60.5 & 238 &  & 1 & 37.8 & 96.2 & 190 \\
     & 0.3 & 82.0 & 85.2 & 168 &  & 0.3 & 42.0 & 136.4 & 133 \\
     & 0.1 & 82.0 & 109.3 & 132 &  & 0.1 & 42.2 & 173.1 & 106 \\
     & 0.03 & 82.0 & 138.1 & 105 &  & 0.03 & 41.4 & 214.8 & 85 \\
     & 0.01 & 82.0 & 168.2 & 86 &  & 0.01 & 41.0 & 256.8 & 70 \\
     & 0.003 & 81.6 & 208.4 & 69 &  & 0.003 & 42.2 & 313.8 & 58 \\
     & 0.001 & 82.3 & 245.3 & 58 &  & 0.001 & 42.2 & 360.9 & 51 \\
     & 0 & 82.2 & 345.5 & 42 &  & 0 & 41.4 & 445.5 & 41 \\
    \bottomrule
\end{tabular}

%% file: tab/llada8b_length512_humaneval_mbpp.tex
\begin{tabular}{ll rrr @{\hspace{1.5em}} ll rrr}
    \toprule
    \multicolumn{5}{c}{HumanEval} & \multicolumn{5}{c}{MBPP} \\
    \cmidrule(lr){1-5}\cmidrule(lr){6-10}
    Method & Param. & Acc. & NFE & TPS & Method & Param. & Acc. & NFE & TPS \\
    \midrule
    OnTrack (Ours) & 0.99 & 40.9 & 419.6 & 34 & OnTrack (Ours) & 0.99 & 50.6 & 222.3 & 38 \\
     & 0.98 & 40.9 & 418.1 & 34 &  & 0.98 & 50.6 & 220.9 & 37 \\
     & 0.96 & 40.9 & 352.6 & 40 &  & 0.96 & 50.6 & 157.0 & 53 \\
     & 0.94 & 40.9 & 285.0 & 49 &  & 0.94 & 50.2 & 124.0 & 68 \\
     & 0.9 & 40.2 & 198.9 & 71 &  & 0.9 & 50.6 & 85.7 & 100 \\
     & 0.85 & 39.6 & 150.2 & 94 &  & 0.85 & 50.2 & 62.4 & 135 \\
     & 0.8 & 37.2 & 124.2 & 113 &  & 0.8 & 49.0 & 54.6 & 158 \\
     & 0.7 & 35.4 & 95.3 & 144 &  & 0.7 & 43.2 & 45.3 & 196 \\
     & 0.6 & 28.0 & 76.3 & 178 &  & 0.6 & 39.3 & 39.4 & 228 \\
     & 0.5 & 25.0 & 60.8 & 215 &  & 0.5 & 31.9 & 35.0 & 254 \\
    \midrule
    Fast-dLLM & 0.99 & 42.1 & 265.3 & 55 & Fast-dLLM & 0.99 & 51.8 & 109.5 & 80 \\
     & 0.98 & 42.1 & 243.1 & 60 &  & 0.98 & 51.8 & 99.9 & 87 \\
     & 0.96 & 42.1 & 215.2 & 68 &  & 0.96 & 51.8 & 87.8 & 99 \\
     & 0.94 & 42.7 & 197.6 & 74 &  & 0.94 & 52.1 & 81.0 & 107 \\
     & 0.9 & 42.7 & 172.5 & 84 &  & 0.9 & 51.8 & 70.9 & 122 \\
     & 0.85 & 43.3 & 151.4 & 95 &  & 0.85 & 52.1 & 64.5 & 134 \\
     & 0.8 & 41.5 & 135.2 & 106 &  & 0.8 & 52.1 & 58.6 & 149 \\
     & 0.7 & 39.6 & 109.0 & 128 &  & 0.7 & 49.8 & 46.9 & 182 \\
     & 0.6 & 34.1 & 86.8 & 159 &  & 0.6 & 47.5 & 43.8 & 209 \\
     & 0.5 & 25.6 & 72.9 & 192 &  & 0.5 & 38.1 & 39.2 & 237 \\
    \midrule
    Learn2PD & 0.99 & 40.2 & 359.7 & 40 & Learn2PD & 0.99 & 51.0 & 177.8 & 47 \\
     & 0.98 & 40.2 & 197.3 & 71 &  & 0.98 & 50.6 & 80.5 & 103 \\
     & 0.96 & 39.6 & 146.3 & 96 &  & 0.96 & 48.6 & 60.9 & 137 \\
     & 0.94 & 37.8 & 124.1 & 112 &  & 0.94 & 48.6 & 53.1 & 158 \\
     & 0.9 & 33.5 & 104.5 & 133 &  & 0.9 & 47.5 & 48.5 & 181 \\
     & 0.85 & 31.7 & 91.0 & 152 &  & 0.85 & 46.7 & 45.4 & 199 \\
     & 0.8 & 29.3 & 79.1 & 169 &  & 0.8 & 43.2 & 42.7 & 217 \\
     & 0.7 & 16.5 & 68.5 & 206 &  & 0.7 & 31.5 & 39.8 & 248 \\
     & 0.6 & 15.9 & 56.2 & 236 &  & 0.6 & 31.9 & 37.7 & 275 \\
     & 0.5 & 12.8 & 46.7 & 289 &  & 0.5 & 21.0 & 35.5 & 303 \\
    \midrule
    EB-Sampler & 10 & 9.8 & 39.0 & 318 & EB-Sampler & 10 & 20.6 & 29.2 & 361 \\
     & 3 & 28.0 & 67.1 & 188 &  & 3 & 37.4 & 43.9 & 225 \\
     & 1 & 37.2 & 108.1 & 129 &  & 1 & 49.8 & 53.2 & 172 \\
     & 0.3 & 41.5 & 160.8 & 88 &  & 0.3 & 52.5 & 70.3 & 123 \\
     & 0.1 & 43.9 & 206.2 & 69 &  & 0.1 & 51.8 & 90.8 & 95 \\
     & 0.03 & 42.1 & 256.5 & 55 &  & 0.03 & 52.1 & 112.0 & 74 \\
     & 0.01 & 42.1 & 298.9 & 47 &  & 0.01 & 51.8 & 137.7 & 62 \\
     & 0.003 & 42.1 & 345.7 & 41 &  & 0.003 & 51.8 & 165.2 & 51 \\
     & 0.001 & 42.1 & 373.8 & 38 &  & 0.001 & 52.1 & 187.7 & 44 \\
     & 0 & 42.1 & 404.7 & 35 &  & 0 & 51.8 & 217.9 & 38 \\
    \bottomrule
\end{tabular}

%% file: tab/llada8b_ci_summary.tex
\begin{tabular}{ll rrr}
\toprule
 & & \multicolumn{3}{c}{NFE speedup (target: \% of reference acc.)} \\
\cmidrule(lr){3-5}
Dataset & Baseline & 90\% & 95\% & 98\% \\
\midrule
GSM8K & Fast-dLLM & \textbf{1.20$\times$ {\scriptsize[1.11, 1.47]}} & \textbf{1.25$\times$ {\scriptsize[1.11, 1.75]}} & \textbf{1.56$\times$ {\scriptsize[1.08, 2.10]}} \\
 & Learn2PD & \textbf{1.10$\times$ {\scriptsize[1.08, 1.20]}} & \textbf{1.40$\times$ {\scriptsize[1.01, 1.65]}} & 1.45$\times$ {\scriptsize[1.00, 5.00]} \\
 & EB-Sampler & \textbf{1.13$\times$ {\scriptsize[1.04, 1.49]}} & \textbf{1.28$\times$ {\scriptsize[1.05, 1.94]}} & 1.59$\times$ {\scriptsize[0.95, 2.52]} \\
\midrule
MATH-500 & Fast-dLLM & \textbf{1.25$\times$ {\scriptsize[1.05, 2.02]}} & 1.50$\times$ {\scriptsize[1.00, 2.02]} & n.r. \\
 & Learn2PD & 1.18$\times$ {\scriptsize[0.98, 1.80]} & 1.38$\times$ {\scriptsize[0.96, 1.84]} & -- \\
 & EB-Sampler & 1.18$\times$ {\scriptsize[0.84, 1.61]} & 1.18$\times$ {\scriptsize[0.84, 1.65]} & n.r. \\
\midrule
HumanEval & Fast-dLLM & 1.15$\times$ {\scriptsize[0.59, 1.46]} & 0.81$\times$ {\scriptsize[0.57, 1.45]} & n.r. \\
 & Learn2PD & 1.23$\times$ {\scriptsize[0.75, 2.69]} & 1.75$\times$ {\scriptsize[0.60, 2.73]} & -- \\
 & EB-Sampler & 1.26$\times$ {\scriptsize[0.59, 1.91]} & 1.00$\times$ {\scriptsize[0.58, 1.94]} & n.r. \\
\midrule
MBPP & Fast-dLLM & 0.95$\times$ {\scriptsize[0.81, 1.26]} & 1.14$\times$ {\scriptsize[0.84, 1.36]} & 1.25$\times$ {\scriptsize[0.83, 1.53]} \\
 & Learn2PD & 0.87$\times$ {\scriptsize[0.83, 1.19]} & 1.05$\times$ {\scriptsize[0.82, 1.33]} & 1.05$\times$ {\scriptsize[0.79, 1.51]} \\
 & EB-Sampler & 1.01$\times$ {\scriptsize[0.95, 1.37]} & 1.01$\times$ {\scriptsize[0.86, 1.40]} & 1.34$\times$ {\scriptsize[0.85, 1.40]} \\
\bottomrule
\end{tabular}